\documentclass[11pt]{article}
\usepackage[final]{acl}

\usepackage{xcolor}

\usepackage[table]{xcolor}
\usepackage{soul}
\definecolor{lightgray}{RGB}{215,215,215}
\definecolor{badRed}{HTML}{B71C1C}
\definecolor{fixBlue}{HTML}{0D47A1}
\definecolor{softGray}{HTML}{F5F5F5}
\newcommand{\badhl}[1]{%
  {\sethlcolor{badRed!20}\textcolor{badRed}{\hl{#1}}}%
}
\newcommand{\fixhl}[1]{%
  {\sethlcolor{fixBlue!18}\textcolor{fixBlue}{\hl{#1}}}%
}
\newcommand{\dimtag}[1]{\textsf{\textbf{#1}}}
\usepackage{capt-of} 
\usepackage{hyperref}
\usepackage{amsfonts}
\usepackage{amsmath}
\usepackage{graphicx}
\usepackage{multirow}
\usepackage{tcolorbox}
\usepackage{adjustbox}
\usepackage{tabularx}
\usepackage{booktabs}
\usepackage{tablefootnote}
\usepackage{booktabs}
\usepackage{array}
\usepackage{balance}
\usepackage{multirow}
\usepackage[normalem]{ulem}
\useunder{\uline}{\ul}{}
\usepackage{subfigure}
\usepackage{algorithm}  
\usepackage{algorithmicx}  
\usepackage[noend]{algpseudocode}  
\usepackage{amsmath}  
\usepackage[inline]{enumitem}
\usepackage{tabularx}
\usepackage[utf8]{inputenc}
\usepackage[english]{babel}
\usepackage{amsthm}
\usepackage{bm}
\usepackage{acronym}
\usepackage{mdframed} 
\usepackage{lipsum}   
\usepackage{times}
\tcbuselibrary{skins,breakable}
\usepackage{listings}
\title{ReportLogic: Evaluating Logical Quality in Deep Research Reports}
\author{Jujia Zhao$^{1,2}$, Zhaoxin Huan$^{2}$, Zihan Wang$^{3}$, Xiaolu Zhang$^{2}$, Jun Zhou$^{2*}$, Suzan Verberne$^{1}$, Zhaochun Ren$^{1*}$ \\
$^1$Leiden Institute of Advanced Computer Science, Leiden University\\
$^2$Ant Group,
$^3$CISPA Helmholtz Center for Information Security\\
\texttt{\{zhao.jujia.0913, zhw.cypher\}@gmail.com}\\
\texttt{\{zhaoxin.hzx, yueyin.zxl, jun.zhoujun\}@antfin.com} \\
\texttt{\{s.verberne, z.ren\}@liacs.leidenuniv.nl}\\
}

\begin{document}
\maketitle
\let\thefootnote\relax\footnotetext{$^*$Corresponding author. This work was supported by Ant Group Research Intern Program.}

\begin{abstract}
Users increasingly rely on Large Language Models (LLMs) for Deep Research, using them to synthesize diverse sources into structured reports that support understanding and action.
In this context, the practical reliability of such reports hinges on \textit{logical quality}: whether the report’s claims and arguments are explicitly supported and can be trusted as a basis for downstream use, rather than merely appearing fluent or informative.
However, current evaluation frameworks largely overlook this requirement.
To bridge this gap, we introduce \textbf{ReportLogic}, a benchmark that quantifies report-level logical quality through a reader-centric lens of \textit{auditability}.
Specifically, ReportLogic adopts a hierarchical taxonomy that evaluates whether readers can
(1) trace an on-topic report structure with a unified analytical arc (\textbf{\textit{Macro-Logic}}),
(2) understand the progression with necessary context (\textbf{\textit{Expositional-Logic}}), and
(3) verify conclusions via explicit claim--support (\textbf{\textit{Structural-Logic}}).
Based on this taxonomy, we construct a human-annotated rubric-guided dataset and train an open-source LogicJudge for scalable evaluation.
We further evaluate judge robustness via adversarial attacks, showing that off-the-shelf LLM judges are frequently influenced by superficial cues (e.g., verbosity), and reasoning modes can mask broken support relations.
Overall, our results provide actionable guidance for building more robust logic evaluators and improving the logical reliability of LLM-generated reports.
\end{abstract}

\section{Introduction}
\label{sec:introduction}


Users increasingly rely on Large Language Models (LLMs) to conduct Deep Research: gathering information from diverse sources, synthesizing it into structured analysis, and producing reports that support understanding and downstream action~\citep{shi2025deep,huang2025deep}.
In this context, the utility of the output report extends beyond producing correct facts or well-formed passages. 
For example, a report may cite accurate statistics and remain well written, yet still be unhelpful when central claims are not adequately supported, leaving readers unable to verify how the conclusion is derived~\citep{mokander2021ethics}.
Consequently, the value of Deep Research hinges on \textit{logical quality}: the capacity to organize information into a unified report-level line of analysis, where the progression of ideas is cohesive and every claim is rigorously grounded.
Without such logical quality, the narrative becomes disjointed and credibility is compromised, fundamentally undermining user trust for downstream use.

\begin{figure}[t]  
\setlength{\abovecaptionskip}{0cm}
\setlength{\belowcaptionskip}{-0.5cm}
    \centering    
    \includegraphics[width=0.95\linewidth]{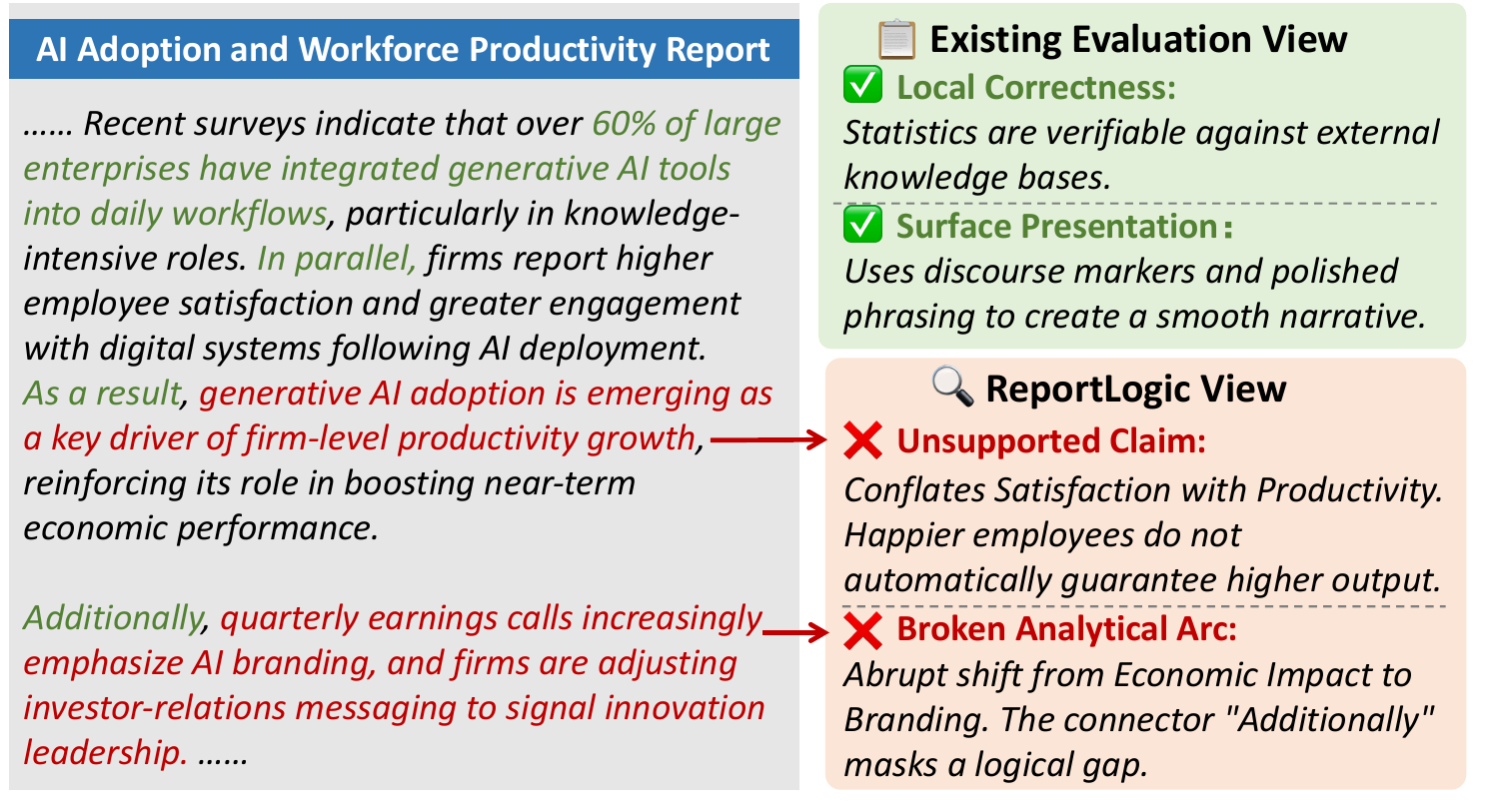}
    \caption{Comparison between existing evaluation views and ReportLogic on a Deep Research report.}
    \label{fig:intro}
\end{figure}

Despite its critical importance, logical quality is still under-specified and weakly measured in current evaluation frameworks for Deep Research and broader long-form generation.
First, current evaluation predominantly prioritizes local correctness (atomic factual accuracy) and surface presentation (fluency, grammar, and discourse markers), focusing on fine-grained factuality checks~\citep{min2023factscore}, verified citations~\citep{sun2024towards}, and transition smoothness~\citep{du2025deepresearch}.
However, these metrics do not assess a report's analytical arc or support relations, i.e., whether the report develops coherently and whether major claims are explicitly supported to justify a well-founded conclusion (Figure~\ref{fig:intro}).
Second, while benchmarks like LogicBench~\citep{parmar2024logicbench} evaluate formal logical reasoning, they focus on formal logic puzzles (e.g., symbolic inference rules) and are not designed to evaluate long-form analytical writing required in Deep Research reports.


Evaluating logical quality in Deep Research poses two fundamental challenges.
(1) Logical quality is inherently multidimensional.
Failures can occur at different granularities, including misalignment with user intent at the organizational level, gaps in discourse that hinder comprehension, or unsupported key claims.
Because these failures operate at distinct scopes, existing holistic scoring often conflates them and provides limited diagnostic feedback.
(2) Logical evaluation lacks a universal decision boundary.
Unlike factuality or citation checks with relatively fixed criteria, judgments of logical sufficiency depend on the specific research question and available context, such as how much evidence is required or which reasoning steps must be explicit.
Together, these challenges motivate a rubric-guided, fine-grained evaluation protocol that translates logical requirements into explicit, instance-specific inspection items.

To bridge the above gap, we introduce \textbf{ReportLogic}, a new evaluation benchmark for assessing logical quality in Deep Research reports.
To enable fine-grained diagnosis across granularities, we move beyond treating logical quality as a monolithic property and operationalize it through a reader-centric lens of \textit{auditability}~\citep{doshi2017towards}: whether a reader can efficiently trace, understand, and verify the analytical process.
Accordingly, ReportLogic adopts a hierarchical taxonomy aligned with these requirements:
(1) \textbf{\textit{Macro-Logic} (Traceability via Organization).}
This layer assesses whether the report stays on topic and assigns clear functional roles to sections, forming a unified analytical arc rather than disjointed points.
(2) \textbf{\textit{Expositional-Logic} (Understandability via Flow).}
Inspired by Information Flow Theory~\citep{halliday2013halliday}, this layer assesses whether the narrative is easy to follow, e.g., by introducing necessary background before abstract analysis and maintaining a coherent Given$\to$New progression.
(3) \textbf{\textit{Structural-Logic} (Verifiability via Argumentation).}
Drawing on the Toulmin model~\citep{toulmin2003uses}, this layer assesses whether key claims are grounded in relevant evidence and whether warrants linking evidence to conclusions are explicit, with appropriate qualifiers.
We further instantiate this taxonomy as eight fine-grained dimensions for diagnostic evaluation.
To make the decision boundary clear for each instance, we implement a context-aware rubric generation protocol that translates each dimension into instance-specific inspection items.

Based on this protocol, we construct a human-annotated benchmark dataset from Deep Research queries drawn from diverse sources, where trained annotators provide dimension-level preferences along with an overall verdict.
To enable scalable evaluation, we further train an open-source \textbf{LogicJudge} to assess logical quality in Deep Research reports and enable community-wide comparison.
We also evaluate judge robustness via adversarial attacks on ReportLogic, finding that off-the-shelf LLM judges are susceptible to superficial cues (e.g., verbosity), and reasoning-optimized judges may mask broken support relations.
Together, these findings provide actionable guidance for developing more resilient logic judges and improving the logical reliability of LLM-generated long-form reports.

Our contributions are threefold:
(1) We introduce ReportLogic, the first human-annotated benchmark for evaluating logical quality in Deep Research, featuring fine-grained dimensions, context-aware instance-specific rubrics, and preference-based annotations.
(2) We propose and train a specialized LogicJudge for scalable evaluation, achieving stronger alignment with human preferences than off-the-shelf LLM judges on ReportLogic.
(3) We perform a systematic adversarial robustness analysis of LLM judges on ReportLogic, uncovering common failure modes and providing fine-grained diagnostic insights into the limitations of current logic judges.\footnote{Our code and data are available at \url{https://github.com/Polaris-JZ/ReportLogic}.}

\section{Related Work}
\label{sec:related_work}

\subsection{Evaluation of Long-form Report Generation}

With advances in long-context modeling~\citep{wu2024longgenbench}, evaluation has moved beyond sentence-level overlap metrics such as BLEU and ROUGE toward benchmarks designed for long-form report generation.
In Deep Research and related long-form reporting scenarios, existing protocols primarily target three content aspects: local correctness, including statement-level factuality~\citep{min2023factscore} and citation verification~\citep{gao2023enabling}; surface presentation, including transition smoothness~\citep{cao2024structeval}; and content coverage, including key-point recall~\citep{wei2024long}.
While effective for correctness and readability, these criteria do not explicitly evaluate report-level logical quality, i.e., whether a report sustains a coherent line of analysis and supports its conclusions through explicit claim--support relations.

\paragraph{Rubric-Guided Evaluation.}
To operationalize the above aspects on open-ended outputs, most recent long-form benchmarks adopt a rubric- or checklist-guided protocol that decomposes holistic quality into a fixed set of sub-judgments.
HelloBench evaluates writing quality and fluency via checklist-style LLM-as-a-judge prompts~\citep{que2024hellobench}; ReportBench grounds atomic factuality in multi-model voting with web-connected LLMs and citation-based verification~\citep{li2025reportbench}; and long-form factuality evaluation relies on atomic claim schemas for fact-checking~\citep{wei2024long, min2023factscore}.
A common property of these protocols is that they apply a \emph{fixed global rubric}, i.e., a shared set of criteria uniformly instantiated across all instances, which offers comparability and reproducibility.
For report-level logical quality, however, this fixed-schema design leaves the decision boundary underspecified on open-domain queries.
In contrast, we adopt an instance-specific rubric conditioned on the query and the compared reports.

\paragraph{LLM-as-Judge and Judge Training.}
Regardless of how rubrics are designed, their practical utility depends on the evaluator that applies them.
Using LLMs as evaluators has become a standard practice for open-ended generation~\citep{kocmi2023large}, but off-the-shelf LLM judges are known to exhibit biases related to length, position, and self-preference~\citep{zheng2023judging}, motivating specialized judge training.
Representative paradigms include supervised fine-tuning on preference data~\citep{kim2023prometheus}, reinforcement learning with judge-specific rewards~\citep{mahan2024generative}, and critique-oriented training that elicits structured rationales before a verdict~\citep{liu2025skywork}.
Most existing trained judges target general-purpose preference or short-form evaluation, rather than long-form, rubric-conditioned logical assessment.
Our LogicJudge is developed for this latter setting: a domain-specialized judge trained on ReportLogic supervision with a rubric-grounded, dimension-wise objective, producing dimension-level diagnostic signals aligned with the benchmark's taxonomy.

\subsection{Logical Evaluation and Argumentation}
Prior work related to logic in NLP can be broadly grouped into formal logical reasoning benchmarks and argumentation mining.
Benchmarks such as LogicBench~\citep{parmar2024logicbench}, FOLIO~\citep{han2024folio}, and JustLogic~\citep{chen2025justlogic} evaluate puzzle-style deductive reasoning in constrained settings with explicit premises and predefined conclusions.
While valuable, they are not designed for open-ended, report-level analytical writing, where quality depends on organization, discourse flow, and sustained claim--support structure across multiple paragraphs.
Argumentation mining studies how argumentative components and relations are expressed in human-written text~\citep{lawrence2019argument}, typically focusing on extracting argument structure at the sentence or argument level~\citep{wachsmuth2017computational}.
It does not typically operationalize logical quality as a multidimensional, report-level property, nor offer an end-to-end evaluation protocol that jointly assesses organization, information flow, and evidential support in generated reports.
\section{The ReportLogic Benchmark}
\label{sec:bench}
To systematically evaluate the logical quality in Deep Research reports, we propose \textbf{ReportLogic}.
Specifically, we focus on \textbf{Deep Research} scenarios, where an LLM synthesizes information from diverse sources into a long-form analytical report in response to an open-domain query.
Each instance comprises (i) a \emph{report-appropriate} query $q$ that requires multi-aspect analysis, and (ii) retrieved contexts $\mathcal{D}=\{d_1,\dots,d_k\}$ returned by a web search pipeline.
Given $(q,\mathcal{D})$, the model generates a report $y$.
ReportLogic evaluates the internal logical quality of the report, rather than external factual correctness or surface writing quality.

\begin{figure*}[t]  
\setlength{\abovecaptionskip}{0cm}
\setlength{\belowcaptionskip}{0cm}
    \centering    
    \includegraphics[width=0.95\linewidth]{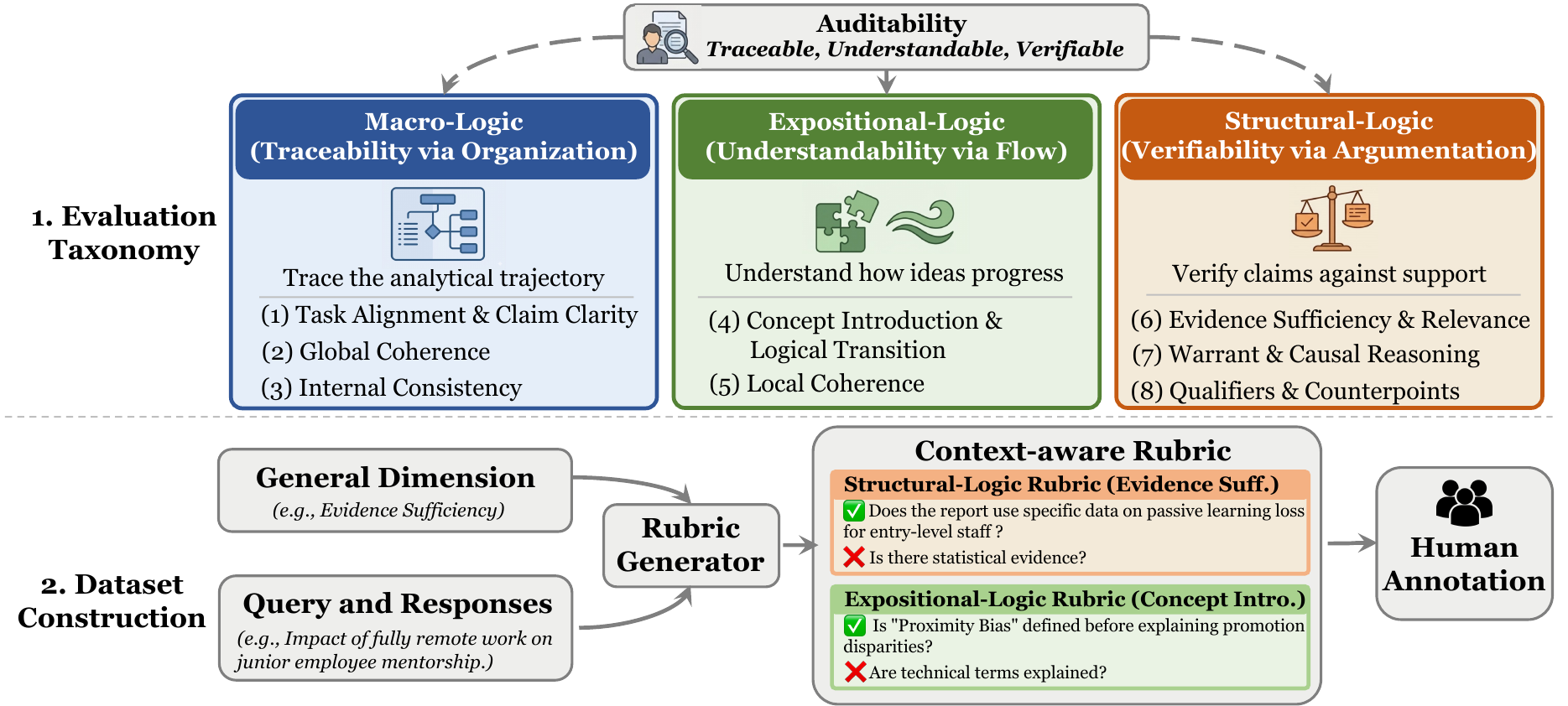}
\caption{ReportLogic framework. 
We define logical quality as auditability and decompose it into a three-layer taxonomy with eight dimensions. Given a query and paired reports, a rubric generator instantiates each dimension into context-aware inspection items that guide pairwise human annotation.}
    \label{fig:method}
\end{figure*}

\subsection{Evaluation Taxonomy}
\label{subsec:taxonomy}
As noted in Section~\ref{sec:introduction}, logical quality in Deep Research is difficult to evaluate because it is inherently multidimensional.
To bridge this, we define logical quality in terms of \emph{auditability}~\citep{doshi2017towards}, namely the extent to which a reader can trace, understand, and verify the analytical process.
We adopt auditability as the organizing principle because Deep Research reports are consumed as decision-support artifacts: a logically reliable report is one whose overall analytical trajectory a reader can \emph{trace}, whose intermediate steps a reader can \emph{understand} without implicit gaps, and whose conclusions a reader can \emph{verify} against explicit reasoning and evidence.
These three reader actions directly induce our top-level taxonomy, providing an actionable grounding for logical evaluation that abstract logic ontologies struggle to supply at the report level.
As Figure~\ref{fig:method} shows, we operationalize auditability through a hierarchical taxonomy that decomposes report-level logic into complementary constraints at different granularities, yielding eight fine-grained dimensions.

\paragraph{Macro-Logic} (Traceability via Report-Level Organization).
Macro-Logic supports auditability at the document scale by providing a clear map of the report’s analytical structure.
A report exhibits strong Macro-Logic if readers can readily trace its central stance and understand how sections contribute to that stance.
This layer includes
(1) \textbf{Task Alignment \& Claim Clarity}, which assesses whether the report commits to a clear central position that directly addresses the query;
(2) \textbf{Global Coherence}, which examines whether sections have well-defined functional roles that support the overall argument; and
(3) \textbf{Internal Consistency}, which checks whether definitions, premises, and quantitative statements remain compatible across the report.

\paragraph{Expositional-Logic} (Understandability via Discourse-Level Explicitness).
Auditability further requires that the progression of ideas remains \textit{understandable}.
A report may possess a coherent high-level organization yet fail logically if it forces the reader to speculate on missing context or bridge implicit gaps between statements.
Guided by Information Flow Theory and the Given-to-New principle~\cite{halliday2013halliday}, Expositional-Logic evaluates discourse-level explicitness, focusing on
(4) \textbf{Concept Introduction \& Logical Transition}, which checks whether new concepts or assumptions are adequately motivated before use, and
(5) \textbf{Local Coherence}, which examines whether adjacent units advance the argument through explicit step-by-step relations rather than rhetorical jumps.

\paragraph{Structural-Logic} (Verifiability via Claim-Support Structure).
Finally, the logical validity of the report depends on the verifiability of the argument itself.
Even a well-structured and easy-to-follow report fails logically if its conclusions rest on weak evidence or hidden inference rules.
Structural-Logic assesses whether major conclusions are grounded in stated support and whether the inferential links from that support to the conclusions are explicit enough to justify the claimed strength.
We ground this layer in Toulmin's model of practical argumentation, which decomposes an argument into Claim, Grounds, and Warrant, optionally with Qualifiers and Rebuttals, providing a concrete schema for verifying completeness and validity of support chains in natural language~\citep{toulmin2003uses}.
It includes
(6) \textbf{Evidence Sufficiency \& Relevance}, which checks whether major claims are supported by specific and probative evidence;
(7) \textbf{Warrants \& Causal Reasoning}, which examines whether inferential bridges are made explicit and causality is used appropriately; and
(8) \textbf{Qualifiers \& Counterpoints}, which evaluates whether conclusions are properly scoped through uncertainty, boundary conditions, or salient alternatives.

\subsection{Dataset Construction}
\label{sec:bench_data}
To instantiate our taxonomy, we construct \textbf{ReportLogic}, a human-annotated dataset for evaluating the logical quality of model-generated Deep Research reports.

\paragraph{Data Collection.}
To obtain sufficient coverage beyond the queries provided in existing Deep Research benchmarks~\citep{xu2025researcherbench,du2025deepresearch}, we additionally source open-domain queries from \textit{Quora} and \textit{Zhihu}\footnote{\textit{Quora} and \textit{Zhihu} are prominent community-driven Q\&A platforms known for hosting high-level, open-ended discussions rather than simple factoid queries.}.
We use GPT-5 to filter and retain only report-appropriate queries, namely prompts that require multi-aspect synthesis and structured analysis rather than short factual replies (Figure~\ref{fig:query_filter} in Appendix~\ref{app:prompts}).
For each retained query, we retrieve supporting context through a web search and passage extraction pipeline, and generate analytical reports using DeepSeek-V3, Claude-4-Sonnet, GPT-4o, and Qwen-Max, with standardized generation prompts in Figure~\ref{fig:gen_prompts} in Appendix~\ref{app:prompts}.
We choose these models as representative systems from different model families and training paradigms, enabling ReportLogic to capture diverse writing and argumentation behaviors under a unified evaluation protocol.

\paragraph{Context-aware Rubric Generation.}
To address the lack of a universal decision boundary described in Section~\ref{sec:introduction}, 
we instantiate our taxonomy into context-aware rubrics.
As illustrated in Figure~\ref{fig:method}, our rubric generator takes as input (i) the target dimension definition, and (ii) the instance context, including the Deep Research query and the paired candidate reports to be compared.
We employ Claude-4.5-Sonnet as the rubric generator to translate each abstract dimension into instance-specific inspection items (Figure~\ref{fig:rubric_prompt} in Appendix~\ref{app:prompts}).
Concretely, for each instance and dimension, it produces (1) a targeted comparison question, (2) span-level cues indicating where to inspect in the reports, and (3) paired good/bad examples that clarify the intended standard in this specific context.
This content-grounded instantiation reduces ambiguity and yields more accurate supervision relative to context-agnostic dimension descriptions, as validated in Section~\ref{subsec:exp_rubric}.
Implementation details of rubric generator selection, annotator-side verification, and schema compliance are reported in Appendix~\ref{app:rubric}.

\paragraph{Human Annotation.}
To obtain reliable preference supervision, we adhere to a multi-annotator adjudication protocol where three trained experts independently evaluate reports using these generated rubrics, with final preference labels determined by majority vote. 
Annotator details are provided in Appendix~\ref{app:human}.
\section{Method: LogicJudge}
\label{sec:judge}
While ReportLogic enables fine-grained human assessment, fully manual evaluation is expensive and slows iteration.
To support scalable benchmarking, model development, and a community-facing leaderboard, we train \textbf{LogicJudge}, an open-source and lightweight judge that predicts pairwise logical preference under our eight-dimensional rubric.
LogicJudge is designed to (i) align with human judgments of logical quality rather than surface cues, (ii) provide actionable, dimension-level diagnostics, and (iii) remain deployable as an automated evaluation component for future research.

\subsection{Task Formulation}
We formulate logical evaluation as a rubric-guided pairwise preference prediction task.
Given a query $q$, context-aware rubrics $\mathcal{R}$, and two candidate reports $(y_A, y_B)$, the judge outputs a structured reasoning chain followed by a final verdict.
The model critiques each of the eight dimensions before predicting the overall preference $d_{\text{all}} \in \{\text{A}>\text{B}, \text{A}<\text{B}, \text{Tie}\}$.
Our label space includes Ties to filter out noise from indistinguishable pairs.
This design mitigates the Halo Effect~\citep{kocmi2023large} by grounding the final verdict in specific logical evidence rather than general impressions or length bias, and provides actionable, fine-grained diagnostics rather than a scalar score.

\subsection{Distilled Training Data}
Constructing a large-scale logic training corpus is challenging due to the high cognitive load of manual annotation.
To overcome this, we synthesize supervision signals via an ensemble distillation protocol using three frontier models: o3, GPT-5, and Gemini 2.5 Pro.
The generation prompts are provided in Figure~\ref{fig:gen_train_prompt} in Appendix~\ref{app:prompts}.
We prioritize label precision over recall through a Dual-Stage Filtering Protocol:
(1) Consensus Filtering: We retain an instance if and only if all three teacher models reach a unanimous verdict. This strict consensus filters out subjective ambiguity and noise.
(2) Swap-Consistency Filtering: For surviving pairs, we swap the order of reports $(y_B, y_A)$ and re-evaluate. We discard instances where the preference does not logically invert.
The final dataset consists of tuples $(q, \mathcal{R}, y_A, y_B, y^*)$, where the reasoning trace $y^*$ is sampled from one of the teacher models.
To facilitate parsing, we constrain the output to a strict JSON schema (schema details in Appendix~\ref{app:schema}).

\subsection{Training Procedure}
We adopt a two-stage alignment curriculum to balance schema validity and discriminative accuracy.
\textbf{(1) Supervised Fine-Tuning (SFT).}
LogicJudge is first initialized via SFT on the distilled training corpus to enforce schema alignment and rubric grounding.
This stage trains the model to produce complete, machine-parseable evaluations aligned with the eight-dimensional taxonomy, rather than generic or surface-level critiques.
\textbf{(2) Group Relative Policy Optimization (GRPO).}
Although SFT ensures stable formatting and rubric adherence, it may not fully capture decision boundaries for hard negative response pairs.
We therefore apply GRPO~\citep{shao2024deepseekmath} with a hierarchical reward to sharpen discriminative preference prediction while preserving strict output structure.
To prevent schema collapse, format validity is enforced as a hard constraint, while label accuracy is treated as a soft incentive.
This \textit{Format-First, Logic-Second} principle enables exploration of reasoning paths without violating the required output schema.
Full objective definitions and reward details are provided in Appendix~\ref{app:training_details}.

\section{Experiments}
\label{sec:exp}

We conduct a comprehensive evaluation to address three questions:
\textbf{RQ1:} Does LogicJudge reliably align with expert judgments of logical quality?
\textbf{RQ2:} How do frontier LLMs perform with ReportLogic, and which logical deficits are most prevalent?
\textbf{RQ3:} How robust are LLM judges to targeted logical perturbations and surface-level bias attacks?

\subsection{Experimental Setup} 
\textbf{Datasets.} 
We use three datasets: DeepResearch (1,204 queries derived from established deep research benchmark~\citep{xu2025researcherbench,du2025deepresearch}), Zhihu (1,262 queries collected from a professional Chinese Q\&A community),\footnote{\url{https://www.zhihu.com/}.} and Quora (1,198 queries sampled from a large English community-driven Q\&A forum).\footnote{\url{https://www.quora.com/}.} 
Dataset statistics are detailed in Appendix~\ref{app:datasets}.

\paragraph{Models and baselines.}
We evaluate LogicJudge against 16 state-of-the-art LLMs across major families, including GPT, Claude, Gemini, Qwen, and DeepSeek, covering both instruction-tuned and reasoning-enhanced variants.
In addition, we include two inference-time ensemble baselines that aggregate the judgments of the same three frontier LLM judges used in our distillation setup: Ensemble Consensus, which outputs a label only when all three judges agree (otherwise abstaining), and Ensemble Vote, which applies majority voting.
All judge baselines are evaluated in a one-shot setting using the same rubric definitions and prompt format as LogicJudge to ensure fairness.
Full baseline configurations are provided in Appendix~\ref{app:baselines}.

\paragraph{Implementation Details.}
We initialize LogicJudge with Qwen-3-30B-A3B\footnote{\url{https://huggingface.co/Qwen/Qwen3-30B-A3B/}.} and train it via a two-stage pipeline (SFT followed by GRPO) using OpenRLHF.\footnote{\url{https://github.com/OpenRLHF/OpenRLHF/}.} 
Hyperparameters and training dynamics are provided in Appendix~\ref{app:setting}.

\begin{table}[t]
\centering
\small
\setlength{\tabcolsep}{6pt}
\renewcommand{\arraystretch}{1.08}
\resizebox{\linewidth}{!}{
\begin{tabular}{lccc}
\toprule
\textbf{Judge Model} & \textbf{DeepResearch} & \textbf{Zhihu} & \textbf{Quora} \\
\midrule

gpt-o3 & 63.73 & 71.70 & 69.67 \\
gpt-4.1 & 68.63 & 70.77 & 34.00 \\
gpt-5 & 62.75 & 71.60 & 70.33 \\
gpt-5.1 & 61.76 & 65.00 & 66.00 \\
\midrule
gemini-2.5-pro & 68.63 & \underline{74.00} & 70.00 \\
gemini-2.5-pro-think & 65.69 & 41.90 & 68.00 \\
gemini-3-pro & 54.90 & 68.20 & 62.33 \\
gemini-3-pro-think & 59.80 & 69.60 & 65.00 \\
\midrule
claude-4-sonnet & 46.08 & 56.70 & 51.67 \\
claude-4-sonnet-think & 45.10 & 56.20 & 51.33 \\
claude-4.5-sonnet & 61.76 & 64.00 & 68.00 \\
claude-4.5-sonnet-think & 63.73 & 64.50 & 66.00 \\
\midrule
qwen3-max & \underline{73.53} & 71.70 & 69.67 \\
qwen-max & 65.69 & 61.62 & 40.67 \\
qwen3-235b & 43.14 & 49.20 & 53.67 \\
\midrule
deepseek-v3 & 62.75 & 66.41 & 53.33 \\

\midrule
Ensemble Consensus & 49.02 & 61.11 & 56.67 \\
Ensemble Vote & 66.67 & 73.57 & \underline{71.31} \\
\midrule

\textbf{LogicJudge (ours)} & \bfseries 74.50 & \bfseries 75.00 & \bfseries 73.00 \\
\bottomrule
\end{tabular}
}
\caption{Judge agreement accuracy (\%) on ReportLogic across three domains. The best and second-best results are highlighted in bold and underlined fonts.}
\label{tab:judge_main}
\end{table}

\subsection{Judge Performance Analysis (RQ1)} 
\label{subsec:exp_judge}
We first assess the reliability of \textbf{LogicJudge} as a surrogate for human experts.
To control for position bias, we report Agreement Accuracy.
For a test pair $(y_A, y_B)$ with ground truth label $l^*$, a prediction is counted as correct only if the model identifies the same winner as the human judge under both the original $(y_A, y_B)$ and swapped $(y_B, y_A)$ orders.
Table~\ref{tab:judge_main} reports results across all three datasets, from which we draw three key findings.
First, LogicJudge achieves the highest agreement with human experts across all domains, outperforming not only strong single-model judges (e.g., GPT-5 and o3) but also the ensemble baselines.
Second, many baselines exhibit large cross-domain variance, with sharp drops on specific datasets, indicating sensitivity to writing style and distribution shift rather than logical quality.
This highlights the need for domain-robust evaluation trained under a fixed logical rubric.
Third, enabling ``thinking'' modes does not consistently improve judgment and can even degrade performance on some domains, suggesting that additional test-time compute introduces noise without task-specific tuning.

\begin{figure*}[t]
\setlength{\abovecaptionskip}{0cm}
\setlength{\belowcaptionskip}{0cm}
    \centering    \includegraphics[width=\linewidth]{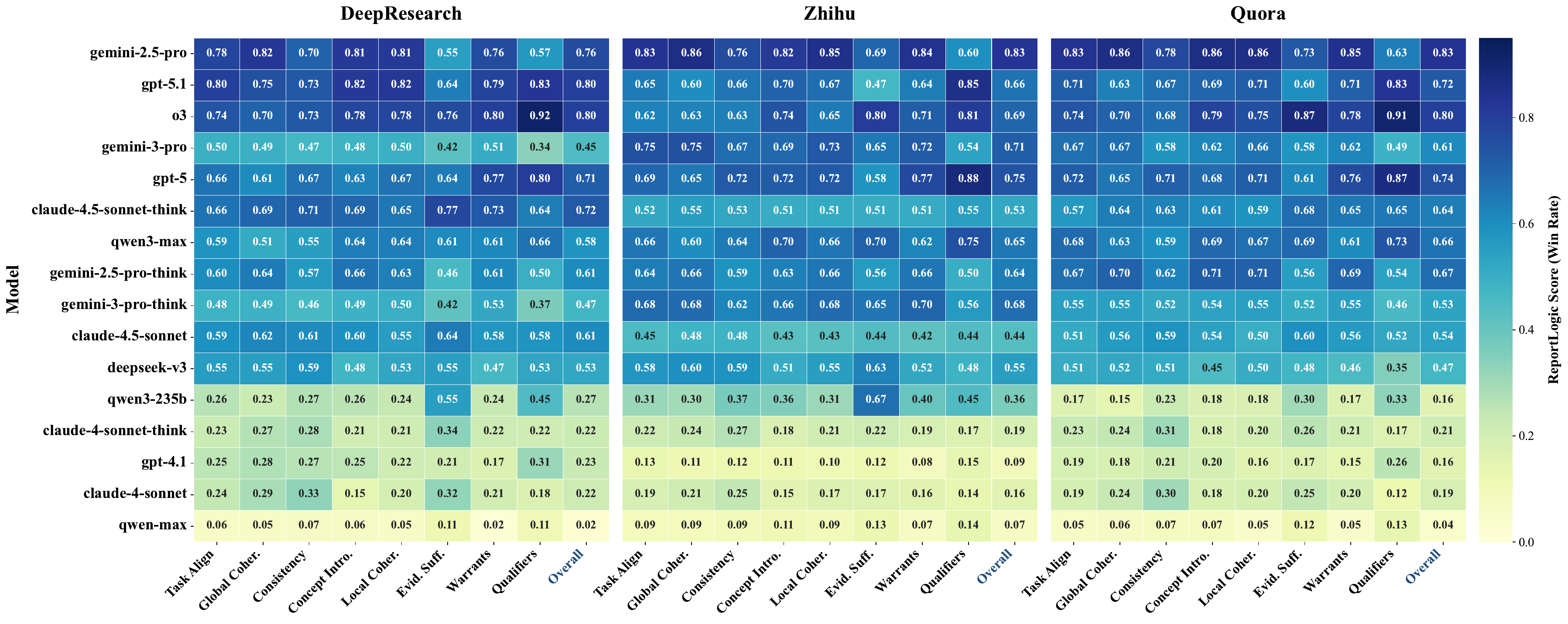}
\caption{ReportLogic Leaderboard. Heatmap of win-rates for 16 frontier models across three domains. Darker colors indicate higher win-rates (stronger logical quality), while lighter colors indicate lower win-rates. Columns are ordered and grouped as follows: the first three dimensions correspond to \textit{Macro-Logic}, the next two to \textit{Expositional-Logic}, the following three to \textit{Structural-Logic}, and the final column reports \textit{Overall}.}
\label{fig:leaderboard}
\end{figure*}

\subsection{Leaderboard Analysis (RQ2)}
\label{subsec:exp_lead}
After validating LogicJudge as a reliable surrogate evaluator, we use it to benchmark the report-level logical quality of 16 state-of-the-art LLMs for analytical report generation.
Figure~\ref{fig:leaderboard} summarizes dimension-wise win rates across three domains.
We highlight three main findings:
(1) Cross-domain stability varies across models.
Gemini-2.5-Pro and GPT-5 exhibit relatively stable performance across DeepResearch, Zhihu, and Quora, indicating that their report-level logic generalizes beyond domain-specific writing styles.
By contrast, models such as Gemini-3-Pro and Claude-4.5-Sonnet shift noticeably across domains, indicating greater sensitivity to dataset-specific conventions and topics.
(2) Longer inference does not reliably improve logical quality.
Reasoning-style (``think'') variants do not consistently outperform their base counterparts and sometimes underperform.
A plausible explanation is that longer reasoning chains can introduce redundancy, topic drift, or over-elaboration that obscures the core argumentative thread, without strengthening rubric-targeted requirements such as clear organization or explicit claim--support links.
(3) Models specialize in different aspects of logical quality.
Specifically, some models perform relatively stronger on \textit{Macro-Logic} and \textit{Expositional-Logic} but are less dominant on \textit{Structural-Logic} (e.g., Gemini-3-Pro, Gemini-2.5-Pro), whereas others show comparatively stronger performance on \textit{Structural-Logic} with weaker or less consistent advantages on \textit{Expositional-Logic} (e.g., o3, GPT-5).
Additional experiment details are provided in Appendix~\ref{app:leaderboard_analysis}.

\begin{figure}[t]
\setlength{\abovecaptionskip}{0cm}
\setlength{\belowcaptionskip}{0cm}
    \centering    \includegraphics[width=\linewidth]{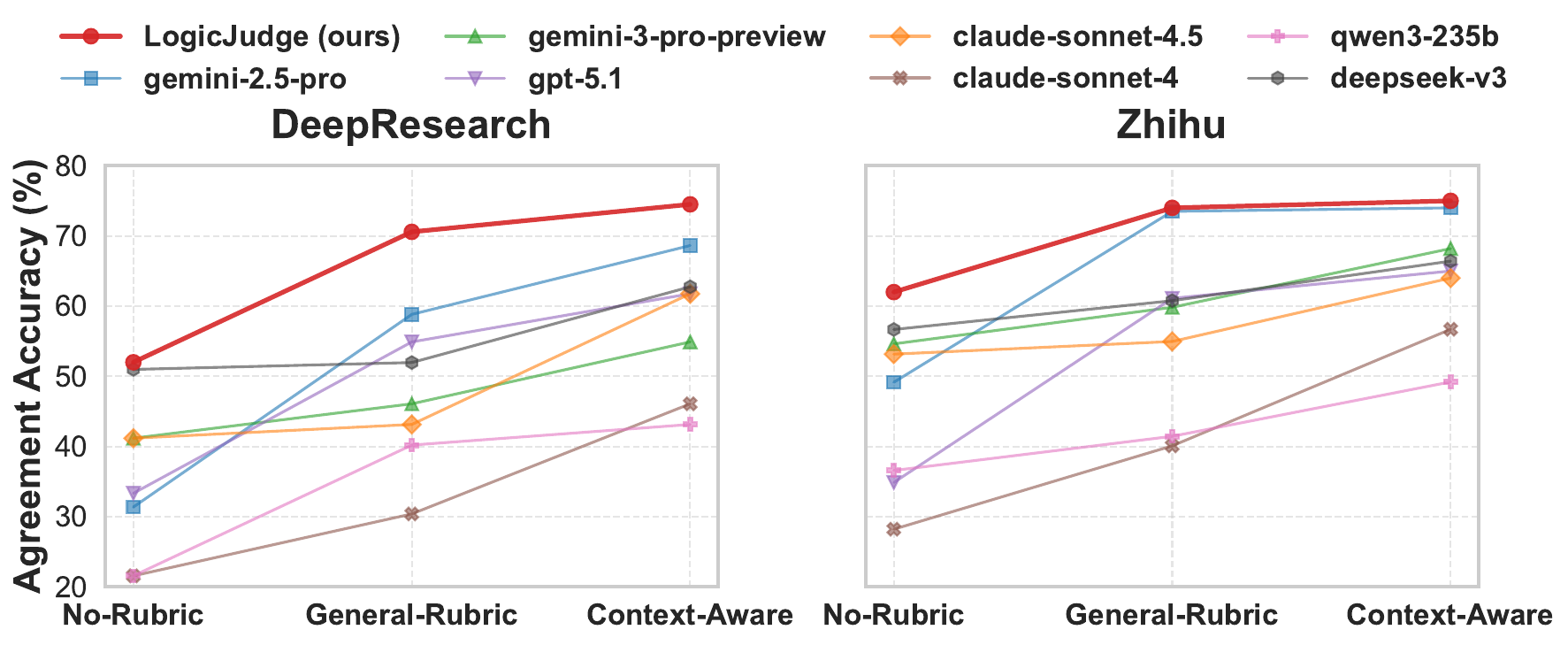}
\caption{Ablation Study on Rubric Effectiveness.}
\label{fig:ablation_rubric}
\end{figure}

\begin{table}[t]
\centering
\small
\resizebox{0.95\linewidth}{!}{
\begin{tabular}{lcc}
\toprule
\textbf{Annotation Setting} & \textbf{Fleiss' Kappa ($\kappa$)} & \textbf{Pairwise Agmt.} \\
\midrule
No-Rubric & 0.37 & 70.11\% \\
General-Rubric & 0.67 & 78.73\% \\
\textbf{Context-Aware } & \textbf{0.71} & \textbf{81.70\%} \\
\bottomrule
\end{tabular}
}
\caption{Human inter-annotator agreement across three settings.}
\label{tab:human_iaa}
\end{table}
\begin{table}[t]
\centering
\small
\begin{tabular}{lcc}
\toprule
\textbf{Dimension} & \textbf{General-Rubric} & \textbf{Context-Aware} \\
\midrule
Task Align.          & 0.68 & 0.71 \\
Global Coher.                          & 0.67 & 0.72 \\
Consistency                      & 0.35 & 0.42 \\
Concept Intro.           & 0.73 & 0.76 \\
Local Coher.                          & 0.64 & 0.68 \\
Evidence Suff.         & 0.74 & 0.78 \\
Warrants              & 0.73 & 0.73 \\
Qualifiers                & 0.67 & 0.74 \\
\bottomrule
\end{tabular}
\caption{Per-dimension human inter-annotator agreement (Fleiss' $\kappa$) under the General-Rubric and Context-Aware settings. The No-Rubric setting does not elicit dimension-level labels and is therefore omitted.}
\label{tab:rubric_agreement}
\end{table}

\subsection{Rubric Effectiveness Analysis (RQ1)}
\label{subsec:exp_rubric}

To assess whether explicit rubric design is necessary for reliable logical evaluation, we study the impact of rubric instantiation on both automated judges and human annotators.
We compare three settings:
(i) \textbf{No-rubric}, which requests a holistic comparison without guidance;
(ii) \textbf{General-rubric}, which provides fixed, context-agnostic definitions for the eight dimensions; and
(iii) \textbf{Context-aware rubric} (ours), which instantiates each dimension into instance-specific inspection items.

We first evaluate multiple LLM judges, including LogicJudge, under all three settings by re-running the same pairwise comparisons on ReportLogic and computing agreement with the human preference labels.
As shown in Figure~\ref{fig:ablation_rubric} (Quora omitted for brevity), rubric instantiation yields a consistent ordering across judges and domains:
context-aware rubric $>$ general rubric $>$ no rubric.
This suggests that high-level dimension definitions alone leave the decision boundary underspecified in long-form logical assessment, while instance-specific inspection items make the comparison criteria operational and more stable.

Additionally, we observe a similar pattern for human annotators. 
As shown in Table~\ref{tab:human_iaa}, rubric guidance substantially increases inter-annotator agreement on the overall preference, with the context-aware rubric producing the most consistent judgments. 
The improvement is not confined to the overall decision: Table~\ref{tab:rubric_agreement} reports per-dimension Fleiss' $\kappa$, showing that context-aware instantiation yields higher or equal agreement across all eight dimensions relative to the general rubric. 

Overall, context-aware rubric instantiation improves reliability by making the decision boundary explicit and enabling more accurate and consistent comparisons for both judges and humans.
Additional experiment details are provided in Appendix~\ref{app:rubric}.

\subsection{Bad Logic Case Analysis (RQ2)}
\label{subsec:exp_bad}
While leaderboard and attack results quantify relative logical performance, they do not reveal how models fail in practice.
To make these failures concrete, we conduct a qualitative analysis of frequently losing reports.
Representative cases, trigger spans, and minimal fixes for each failure mode are provided in Appendix~\ref{app:bad_cases}.
Across datasets and models, we identify three recurring failure patterns.
First, models violate global consistency constraints by producing locally plausible claims that cannot jointly hold at the document level.
Second, reports often exhibit smooth but weakly connected discourse, relying on rhetorical transitions rather than explicit inferential links.
Third, conclusions are frequently supported by implicit or missing warrants, with asserted causes lacking specified mechanisms.

\begin{figure}[t]  
\setlength{\abovecaptionskip}{0cm}
\setlength{\belowcaptionskip}{0cm}
    \centering    
    \includegraphics[width=0.95\linewidth]{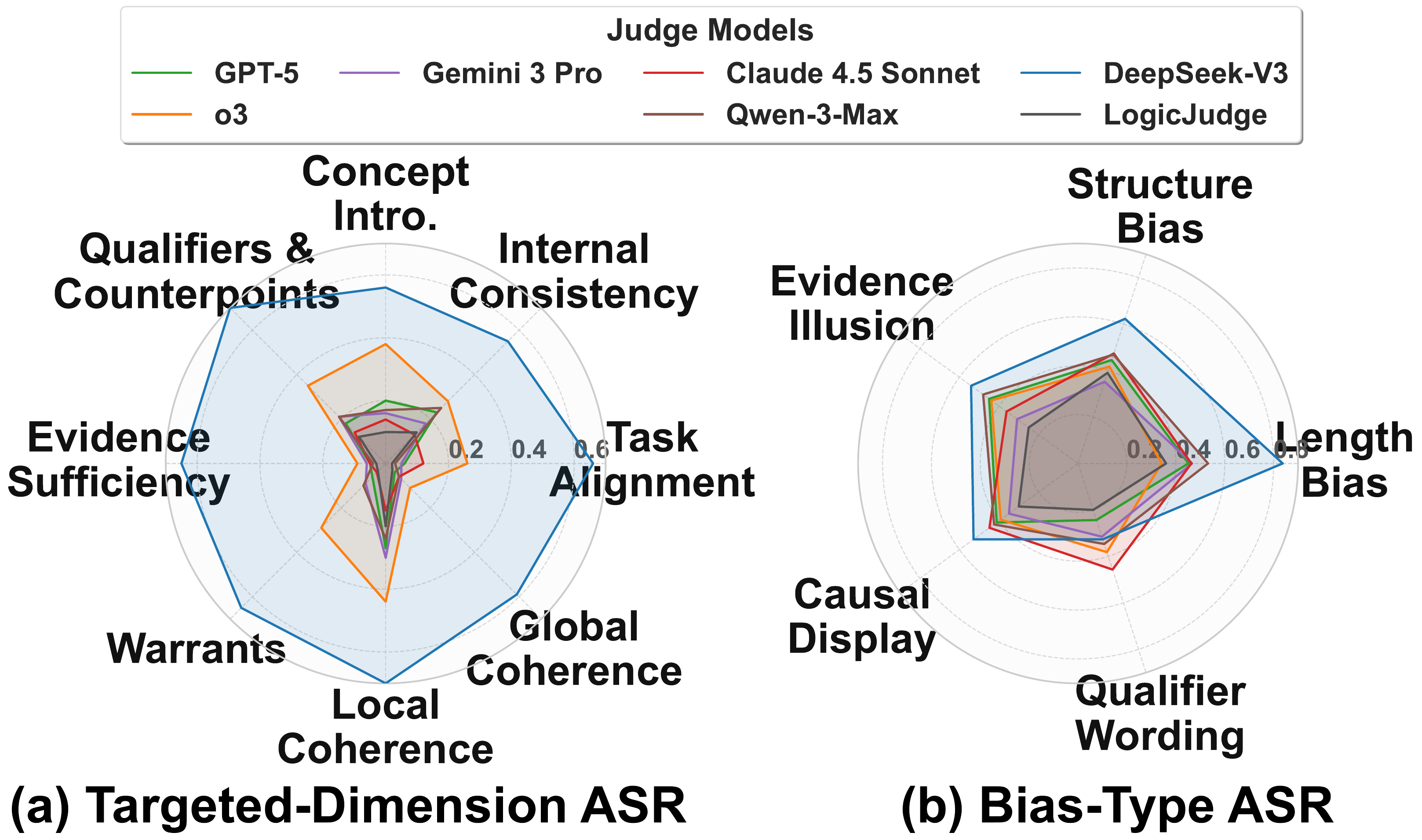}
  \caption{Attack analysis of judge robustness.
(a) Targeted-dimension Attack Success Rate (ASR): fraction of cases where the judge prefers the attacked response, with lower indicating better robustness.
(b) Bias-type ASR: fraction of cases where the judge prefers a logically equivalent response with surface manipulations.}
  \label{fig:target_dim_attack}
\end{figure}

\subsection{Attack Analysis (RQ3)}
\label{subsec:exp_attack}
To assess whether LLM judges can distinguish genuine logical defects from superficial biases and evaluate their robustness under adversarial perturbations, we conduct an attack analysis on ReportLogic.
We apply adversarial attacks to $N=300$ sampled instances using two complementary attack types:
(i) a \textbf{Targeted-Dimension Attack}, which injects a localized defect into a specific rubric dimension; and
(ii) a \textbf{Bias-Type Attack}, which preserves the underlying logical content while manipulating surface cues such as verbosity or formatting.
We report Attack Success Rate (ASR) to measure vulnerability.
Results are shown in Figure~\ref{fig:target_dim_attack}.

First, sensitivity to targeted logical defects varies substantially across judges.
As shown in Figure~\ref{fig:target_dim_attack}(a), DeepSeek-V3 is the most vulnerable, exhibiting high ASR across nearly all dimensions, while GPT-5, Gemini 3 Pro, Claude 4.5 Sonnet, and Qwen-3-Max show substantially lower ASR.
Notably, o3 exhibits elevated ASR on both \textit{Warrants} and \textit{Local Coherence}, suggesting that reasoning-optimized judges tend to actively repair perceived gaps by inferring implicit warrants and smoothing over local incoherence, which can mask genuine logical defects.
Second, surface-level biases can substantially distort judge preferences under the bias-type attack.
Figure~\ref{fig:target_dim_attack}(b) shows that \textit{Length Bias} is the most consistently effective manipulation, and \textit{Structure Bias} and \textit{Evidence Illusion} also produce non-trivial vulnerability.
In contrast, \textit{Qualifier Wording} exhibits the weakest effect overall.
LogicJudge is relatively more robust overall, exhibiting consistently low ASR across most dimensions.
Additional attack details and analyses are provided in Appendix~\ref{app:attack}.

\section{Conclusion}
\label{sec:conclusion}

In this work, we argued that the reliability of Deep Research reports hinged not only on factual correctness or fluency, but on their \emph{logical quality}: whether claims were coherently organized, explicitly supported, and verifiable by readers.
Specifically, we introduced ReportLogic, a benchmark that operationalizes logical quality through a reader-centric notion of auditability.
By decomposing logic into a hierarchical taxonomy and instantiating it via context-aware rubrics, ReportLogic enables fine-grained, diagnostic evaluation of analytical reports.
Our human annotations, LogicJudge, and adversarial attack show that explicit rubric guidance is critical for reliable logical assessment, and that off-the-shelf LLM judges remain vulnerable to superficial cues and implicit reasoning shortcuts.
Improving the logical quality of LLM-generated reports is essential for building user trust and enabling effective real-world use.
We hope ReportLogic provides a foundation for logic-aware evaluators and principled evaluation of Deep Research reports.

\clearpage

\section*{Limitations}
\label{app:limitations}
First, ReportLogic is designed for Deep Research-style analytical report generation, where logical quality is operationalized from the view of auditability.
Other genres such as creative writing, narrative storytelling, or purely persuasive rhetoric may follow different conventions and trade-offs for what constitutes ``good logic,'' and adapting our taxonomy and rubrics to those settings requires further validation and redesign.
Second, while our benchmark and judge identify logical deficits in Deep Research reports, we do not directly address how to improve logical generation itself.
Retrieval and planning offer promising opportunities to enforce a stronger global structure and verifiable support through evidence coverage, argumentative organization, and explicit intermediate outlines.


\section*{Ethical Considerations and Artifacts.}
This work uses publicly available datasets and model-generated reports, and all third-party resources are cited and used under their original research terms. We release ReportLogic and LogicJudge solely for research and diagnostic evaluation of long-form logical quality, not for deployment in downstream decision-making or moderation. We do not collect or model personally identifiable information, and we report results only in aggregate. We filter and remove instances containing offensive content during curation and annotation.
\clearpage

\bibliography{reference}
\clearpage
\appendix
\section*{Appendix}

\section{LogicJudge Training}
\subsection{Output Schema for LogicJudge}
\label{app:schema}
To enforce comprehensive rubric coverage and enable reliable automated parsing, we constrain the output of LogicJudge to follow a strict, machine-verifiable schema.
The training prompt used to elicit this format is provided in Figure~\ref{fig:logic_pref_prompt}.
Each prediction consists of two components: a structured reasoning block enclosed in a \texttt{<think>} tag, followed by a single final preference token.

\paragraph{Structured Reasoning Block.}
The \texttt{<think>} block contains a JSON object that explicitly records the judge's intermediate, dimension-level evaluations.
This object must include the following two top-level fields:

\begin{itemize}
    \item \texttt{aspect\_evaluations}: a dictionary mapping each of the eight logical dimensions to a discrete comparison decision and a brief justification.
    \item \texttt{overall\_explanation}: a short natural-language explanation summarizing the primary factors driving the overall preference.
\end{itemize}

Each entry in \texttt{aspect\_evaluations} corresponds to one logical dimension (Section~\ref{subsec:taxonomy}) and follows a uniform structure with two required fields:
\begin{itemize}
    \item \texttt{decision}: a categorical label from the set \{\texttt{A>B}, \texttt{A<B}, \texttt{both\_good}, \texttt{both\_bad}\}, indicating the relative logical quality under that dimension.
    \item \texttt{justification}: a concise textual rationale grounded in the content of the two reports and the instantiated rubric.
\end{itemize}

\paragraph{Final Decision Token.}
After the \texttt{<think>} block, the model outputs exactly one deterministic token representing the overall preference label (e.g., \texttt{A>B}).
This separation allows downstream systems to parse the final verdict efficiently while retaining access to dimension-level diagnostics when needed.

\paragraph{Schema Illustration.}
An example of a valid output structure is shown below:

\begin{lstlisting}
<think>
{
  "aspect_evaluations": {
    "task_alignment_claim_clarity": {
      "decision": "Tie",
      "justification": "Both reports directly answer the query by attributing Transformer dominance to architectural advantages over RNN/LSTM, and both keep this thesis stable throughout."
    },
    "global_coherence": {
      "decision": "A<B",
      "justification": "B cleanly organizes the answer into a criteria-driven structure (parallelism, long-range dependency, architectural flexibility, multi-head attention), making the argument easier to follow at the document level."
    },
    "...": { }
  },
  "overall_explanation": "On the query of why Transformers overtook RNN/LSTMs, Report B is more convincing because it systematically decomposes the advantages and backs them with concrete benchmarks and deeper mechanism explanations (e.g., constant path length mitigating long-range issues). A is smoother and more balanced, but B's structure-plus-evidence makes the core argument stronger overall."
}
</think>
A<B
\end{lstlisting}

\paragraph{Design Rationale.}
This schema serves two purposes.
First, by requiring explicit evaluations for all eight dimensions, it prevents the judge from collapsing into holistic or surface-level judgments.
Second, the strict separation between structured reasoning and the final decision ensures robustness for both training and deployment: unparsable outputs are immediately identifiable, and valid outputs provide fine-grained diagnostic signals in addition to a scalar preference.

\subsection{Training Details for LogicJudge}
\label{app:training_details}

This appendix provides the full objective definitions, reward design, and optimization details for training the LogicJudge model, complementing the high-level description in Section~\ref{sec:judge}.

\subsubsection{Supervised Fine-Tuning Objective}
The primary objective of SFT is to adapt the model to the specific rubric-following and schema-compliance requirements of ReportLogic. 
Given the dataset of curated tuples $D = \{(x, y^*)\}$, where $x=(q, \mathcal{R}, y_A, y_B)$ is the input and $y^*$ is the label, we optimize the standard autoregressive object:
\begin{equation}
    \mathcal{L}_{SFT}(\theta) = -\mathbb{E}_{(x, y^*) \sim D} \left[ \sum_{t=1}^{T} \log \pi_\theta(y^*_t \mid y^*_{<t}, x) \right]
\end{equation}
SFT serves as a critical initialization for two reasons:
(1) Schema Alignment: It teaches the model to adhere to the rigid format required for automated parsing.
(2) Rubric Grounding: It aligns the model's natural language explanations with our eight-dimensional taxonomy, preventing the generation of generic or irrelevant critiques.




\subsubsection{Group Relative Policy Optimization}
While SFT establishes formatting and basic reasoning, it may not fully capture the decision boundaries for hard negatives. 
To further optimize discriminative performance, we employ Group Relative Policy Optimization (GRPO)~\citep{shao2024deepseekmath}.

Unlike standard Proximal Policy Optimization (PPO), which relies on a separate value network (critic) to estimate the baseline and often introduces significant memory overhead, GRPO estimates the baseline directly from the group statistics of sampled outputs.
Formally, for each input $x$, we sample a group of $G$ outputs $\{o_1, \dots, o_G\}$ from the current policy $\pi_{\theta_{old}}$, and let the probability ratio be $r_i(\theta) = \frac{\pi_\theta(o_i|q)}{\pi_{\theta_{old}}(o_i|q)}$. The GRPO objective is:
\begin{equation}
\resizebox{0.99\linewidth}{!}{$
    \mathcal{J}_{GRPO}(\theta) = \mathbb{E}_{q, \{o_i\}} \left[ \frac{1}{G} \sum_{i=1}^G \left( \mathcal{M}_i(\theta) - \beta \mathbb{D}_{KL} \right) \right]$}
\end{equation}
where $\mathcal{M}_i(\theta)$ represents the clipped surrogate objective, with $\epsilon$ denoting the clipping threshold that limits the deviation of the policy update:
\begin{equation}
    \min \left( r_i(\theta) \hat{A}_i, \text{clip}\left(r_i(\theta), 1-\epsilon, 1+\epsilon\right) \hat{A}_i \right)
\end{equation}
where the advantage $\hat{A}_i$ is computed by normalizing the rewards within the group: $\hat{A}_i = \frac{r_i - \text{mean}(\{r_j\})}{\text{std}(\{r_j\})}$. 
This group-based normalization is particularly effective for reasoning tasks, as it robustly distinguishes better reasoning paths from worse ones within the same context, independent of the absolute difficulty of the input query.

A key challenge in RL for structured generation is structure collapse, where the model optimizes for label correctness but forgets the output schema.
To prevent this, we design a hierarchical reward function that treats format validity as a hard prerequisite. 
For a generated output $o$, the total reward $r(o)$ is computed as:
\begin{equation}
    r(o) = 
    \begin{cases} 
    r_{format} & \text{if format is invalid} \\
    r_{acc} & \text{if format is valid}
    \end{cases}
\end{equation}

\begin{enumerate}
    \item \textbf{Hard Format Constraint ($r_{format}$):} We parse the output to verify it is a valid JSON object containing all eight required dimension keys and a final decision token. If parsing fails, the model receives a severe penalty (e.g., $r_{format} = -1$). This ensures that structural validity is prioritized above all else, as an unparsable report is useless to the pipeline.
    \item \textbf{Soft Accuracy Incentive ($r_{acc}$):} If and only if the format is valid, we evaluate the correctness of the final decision label against the consensus ground truth. Correct predictions yield a positive reward (e.g., $+1$), while incorrect ones yield a mild penalty (e.g., $-0.5$).
\end{enumerate}
This ``Format-First, Logic-Second'' shaping stabilizes training, ensuring the judge remains deployable while iteratively improving its reasoning accuracy.

\section{Implementation Details}

\subsection{Dataset Details}
\label{app:datasets}
\begin{table*}[t]
\centering
\small
\setlength{\tabcolsep}{6pt}
\renewcommand{\arraystretch}{1.15}
\begin{tabular}{lccccc}
\toprule
\textbf{Dataset} &
\textbf{\#Queries} &
\textbf{Train/Val/Test (queries)} &
\textbf{\#Pairs (human-labeled)} &
\textbf{\#Pairs (distilled)} \\
\midrule
DeepResearch 
& 1204
& 1055 / 17 / 17 
& 102 
& 7912 / 94 \\
Zhihu        
& 1262  
& 843 / 224 / 195   
& 1170 
& 15210 / 1424 \\
Quora        
& 1198  
& 1098 / 50 / 50   
& 300 
& 6590 / 336 \\
\midrule
\textbf{Total} 
& 3779 
& 2996 / 291 / 262 
& 1572 
& 29712 / 1854 \\
\bottomrule
\end{tabular}
\caption{Dataset statistics. All splits are performed at the query level.
Human-labeled pairs appear only in the test split and are annotated by three experts with majority voting.
Distilled pairs are used for LogicJudge training and validation after consensus and swap-consistency filtering.}
\label{tab:dataset_stats}
\end{table*}

We summarize the construction of the datasets used in our experiments, including query sourcing, Deep Research suitability filtering, query-level splitting, and pairwise expansion.
Table~\ref{tab:dataset_stats} reports the resulting statistics for evaluation and judge training.

\paragraph{Query sources and filtering.}
We collect open-domain candidate queries from three sources: (i) existing Deep Research benchmarks~\citep{xu2025researcherbench,du2025deepresearch}, (ii) Zhihu, a professional Chinese community-driven Q\&A platform, and (iii) Quora, a large English community-driven Q\&A forum.
To ensure that retained queries are genuinely report-appropriate, we apply the Deep Research suitability filtering protocol described in Section~\ref{sec:bench} (Figure~\ref{fig:query_filter}), retaining only queries that require retrieval-dependent, multi-aspect synthesis and structured long-form analysis.
Queries violating hard constraints (e.g., political content or closed-form tasks) are discarded.

\paragraph{DeepResearch domain and synthetic expansion.}
The DeepResearch Benchmark is intrinsically small, containing only 102 original queries~\citep{xu2025researcherbench,du2025deepresearch}.
Given the high cost of expert human annotation and the need to allocate original queries across training, validation, and testing, it is not feasible to reserve the entire set for human evaluation.
To increase coverage during LogicJudge training, we therefore expand only the training split with synthetic query variants derived from the original DeepResearch training queries, which provide additional topical and linguistic diversity.
Synthetic variants are produced by controlled rewriting of the original training queries and are subjected to near-duplicate filtering to avoid trivial overlap.
These synthetic queries are used exclusively for training and do not define the evaluation distribution.
Importantly, all human evaluation on the DeepResearch domain is conducted on held-out original (non-synthetic) queries, ensuring that test results reflect the authentic query distribution of the benchmark.

\paragraph{Query-level split and pairwise expansion.}
We split each dataset at the query level into train/validation/test, ensuring no query overlaps across splits.
For each query, we generate multiple candidate reports using the set of report-generation models described in Section~\ref{sec:bench}.
We then construct pairwise comparison instances by pairing reports generated for the same query.

\paragraph{Human annotation vs.\ distilled supervision.}
Because evaluating long-form logical quality is cognitively demanding, we reserve human annotation for the test split only.
Each test pair is labeled by three trained experts with majority vote.
In contrast, the train/validation splits are used exclusively to train and tune LogicJudge and are constructed via the distillation protocol in Section~\ref{sec:judge}, which applies consensus filtering and swap-consistency filtering to retain high-confidence preference signals.

\subsection{Models and Baselines}
\label{app:baselines}

We detail the models used as judge baselines in our experiments, including model families, variants, and evaluation settings.

\paragraph{Model Families.}
We evaluate \textbf{LogicJudge} against 16 state-of-the-art LLMs spanning both proprietary and open-weight ecosystems.
The selected baselines cover five major model families:

\begin{itemize}[leftmargin=*]
    \item \textbf{GPT Series.}
    We include GPT-4.1, GPT-5, and GPT-5.1 as representative instruction-tuned models, as well as o3, a frontier reasoning-oriented model designed for complex multi-step inference.

    \item \textbf{Claude Series.}
    We evaluate Claude-3.5-Sonnet and Claude-4-Sonnet, together with their reasoning-enhanced variants (Sonnet-3.5-Think and Sonnet-4-Think).
    These variants explicitly increase test-time computation, allowing us to isolate the effect of extended reasoning on logical judgment.

    \item \textbf{Gemini Series.}
    We include Gemini-2.5-Pro and Gemini-3-Pro, along with their corresponding thinking-mode variants.
    Similar to Claude, these models enable inference-time scaling for deeper chain-of-thought reasoning.

    \item \textbf{Qwen Family.}
    To represent strong open-weight models, we evaluate Qwen-Max, Qwen3-Max, and Qwen3-235B.
    These models are competitive instruction-tuned systems widely used in open research settings.

    \item \textbf{DeepSeek.}
    We include DeepSeek-V3, a high-performing open-weight model with strong generation capabilities, which serves as a representative baseline for non-specialized judges.
\end{itemize}

\paragraph{Reasoning-Enhanced Variants.}
For models that provide explicit ``thinking'' or reasoning modes (e.g., Claude-Think, Gemini-Think), we evaluate both the standard and reasoning-enhanced versions.
These variants are designed to allocate additional inference-time compute to internal reasoning processes.
Evaluating both variants allows us to assess whether increased test-time reasoning alone improves logical judgment, independent of task-specific alignment.

\subsection{Training Settings and Hyperparameters}
\label{app:setting}

We provide the detailed training configuration for LogicJudge, complementing the high-level description in Section~\ref{sec:judge}.
All training runs were conducted on NVIDIA A100 80GB GPUs.

\paragraph{Stage 1: Supervised Fine-Tuning (SFT).}
The SFT stage aligns the model with the rubric-guided reasoning format and the structured output schema required for automated parsing.
We fine-tune the model on the distilled training split for 5 epochs using a global batch size of 128 and a learning rate of $1\times10^{-5}$.
To support the long contexts required by Deep Research reports, the maximum input sequence length is set to 40{,}960 tokens.
SFT is trained using 16 A100 GPUs and takes approximately 6 hours.
To prevent overfitting and to provide a stable initialization for reinforcement learning, we select the checkpoint with the lowest validation loss on the distilled validation set as the starting point for the subsequent GRPO stage.

\paragraph{Stage 2: Group Relative Policy Optimization (GRPO).}
After SFT convergence, we further refine LogicJudge using GRPO to sharpen discriminative preference judgments on challenging report pairs.
During GRPO training, we use a large global batch size of 2{,}048 and a rollout batch size of 512 to ensure reliable estimation of group-relative advantages.
The initial KL coefficient is set to 0.001 to constrain policy drift, and the sampling temperature is fixed at 1.0 to balance exploration and stability.
The maximum generation length for the judge’s reasoning output is capped at 4{,}096 tokens.
GRPO is trained using 32 A100 GPUs and takes approximately 16 hours.

\paragraph{Evaluation Protocol.}
All baseline models are evaluated strictly as judges, not generators.
Each model receives the same input format consisting of:
(i) the user query,
(ii) the instantiated context-aware rubric,
and (iii) a pair of candidate reports $(y_A, y_B)$.

To ensure fairness and isolate judgment capability:
(1) All baselines are evaluated in a one-shot setting without any additional demonstrations.
(2) The exact same prompt template and rubric definitions used for LogicJudge are applied to all baselines, as shown in Figure~\ref{fig:logic_pref_prompt}.
(3) No model-specific prompt engineering or output post-processing is used.

Models are required to output a pairwise preference decision, which is evaluated using the agreement-based metrics described in Section~\ref{subsec:exp_judge}.

\paragraph{Deterministic Parsing Rule.}
Some baseline models do not reliably follow the structured output schema required by LogicJudge.
For such models, we extract the final pairwise preference using a deterministic parsing rule.
Specifically, we identify the model’s explicit comparative judgment between the two responses (e.g., “A is better than B”, “B outperforms A”, or tie statements such as “both are equally good”), prioritizing the final explicit decision when multiple judgments are present.
If no unambiguous preference can be identified, the instance is excluded from agreement-based evaluation.
This procedure is fully rule-based and does not introduce additional learned components or heuristics.

\section{Experiment Analysis}

\subsection{Leaderboard Analysis}
\label{app:leaderboard_analysis}
We provide additional details and analyses for the leaderboard in Figure~\ref{fig:leaderboard}, including the evaluation setting, how win-rates are computed.

For each domain, we use the full set of test queries in ReportLogic.
For every query, we prompt each of the 16 evaluation models to generate a report using the same prompt shown in Figure~\ref{fig:gen_prompts}, resulting in one report per model per query.
We then construct pairwise comparisons by matching reports produced for the same query across all model pairs.
For each comparison pair, we apply the same context-aware rubric generation protocol as in Section~\ref{sec:bench_data} to produce instance-specific inspection items under our taxonomy, conditioned on the query and the two compared reports.
Given these rubrics, LogicJudge performs a dimension-wise pairwise judgment and selects the preferred report for each dimension.
A model receives a win on a dimension whenever its report is preferred over its paired opponent for the same query.
The win-rate on a given domain and dimension is computed as the fraction of pairwise comparisons a model wins, averaged over all queries and opponents; ties are excluded.
Figure~\ref{fig:leaderboard} visualizes the resulting dimension-wise win-rates as a heatmap.

\subsection{Rubric Effectiveness}
\label{app:rubric}
We provide full experiment setting details of the rubric ablation summarized in Section~\ref{subsec:exp_rubric} here, together with a discussion of rubric-side quality assurance that underpins the ablation.

\paragraph{Rubric Quality Assurance.}
Beyond the downstream effectiveness analysis, we also ensure rubric quality at the generation stage through model selection and annotator-side verification. During development, we compared several candidate rubric generators, including Gemini-2.5-Pro, GPT-5.1, and Claude-4.5-Sonnet. For each candidate, we manually inspected sampled rubrics for schema compliance, semantic alignment with the target dimension, and usability during annotation. We selected Claude-4.5-Sonnet because it consistently produced clearer, more structured, and better dimension-aligned rubrics than the alternatives. During annotation, generated rubrics are not treated as unquestioned ground truth: each annotator first verifies that the rubric is well-formed and usable (i.e., the guiding question targets the intended dimension, the good/bad examples are consistent with that question, and the span hints point to inspectable regions) before applying it to report comparison. Ill-formed rubrics are flagged and regenerated. Across the full dataset, we observed no violations of the required JSON schema, which we attribute to constrained prompting and the strong instruction-following behavior of the selected generator.

\paragraph{Ablation Protocol.}
We compare three conditions that differ only in how decision boundaries are specified:
No-Rubric (holistic comparison without explicit criteria),
General-Rubric (fixed, context-agnostic definitions of the eight dimensions), and
Context-Aware Rubric (ours; instance-specific inspection targets including a comparison question, span-level cues, and paired good/bad examples for each dimension).
For this ablation, we randomly sample 200 instances from the ReportLogic test set.
All other factors are controlled (same report pairs, same judge prompt template aside from the rubric content, and the same evaluation metric).

\paragraph{Judge-side experiment setting.}
We evaluate multiple judge models (including LogicJudge and selected frontier LLM judges) on all three datasets.
We use the same agreement-based metric as the main experiments: a prediction is counted as correct only if the judge selects the same winner when the pair is presented in the original order and when the two responses are swapped, controlling for position bias.

\paragraph{Human-side experiment setting.}
We conduct a controlled annotation study with three trained expert annotators under the same three rubric conditions.
We report inter-annotator agreement using Fleiss' Kappa ($\kappa$) and average pairwise agreement.
Specifically, for each report pair, each annotator provides a categorical preference label in $\{\text{A}>\text{B}, \text{A}<\text{B}, \text{Tie}\}$.
Average pairwise agreement is computed as the mean, over all report pairs, of the fraction of annotator pairs that assign the same label; with three annotators, this equals the average of the three pairwise match indicators.
Fleiss' Kappa $\kappa$ measures inter-annotator agreement beyond chance for multiple raters.
It compares the observed agreement among annotators with the agreement that would be expected if annotators made decisions according to the overall label distribution at random.
Formally, $\kappa = \frac{\bar{P}-\bar{P}_e}{1-\bar{P}_e}$, where $\bar{P}$ denotes the average observed agreement across report pairs, and $\bar{P}_e$ denotes the expected agreement under random labeling based on empirical label frequencies.

\subsection{Qualitative Bad-Case Analysis}
\label{app:bad_cases}
While the leaderboard provides a quantitative ranking of relative logical quality, it does not reveal the concrete mechanisms behind model failures in Deep Research-style report generation. 
To make these deficits visible and actionable, we conduct a qualitative bad-case analysis on frequently losing reports in Figure~\ref{tab:bad_cases}.

\noindent\textbf{(1) Internal Inconsistency.}
Internal consistency requires a report to satisfy global constraints and avoid contradictions across sections, which is a prerequisite for auditable long-form argumentation.
However, we observe that models can produce locally plausible details that violate global constraints.
In Case~1 (Figure~\ref{tab:bad_cases}), when reporting ``Chinese Social Stratification,'' the model assigns population shares to multiple strata, yet the resulting quantities cannot simultaneously hold: the implied totals (e.g., 45\% for Class~4 and $\sim$25\% for Classes~6--9, in addition to other stated groups) exceed 100\%. 
This pattern indicates that numerical statements are often generated as individually reasonable fragments without being reconciled under shared global constraints, undermining the report's verifiability.

\noindent\textbf{(2) Local Incoherence.}
We find that many reports maintain surface smoothness via generic discourse markers (e.g., ``however,'' ``besides,'' ``from a career perspective''), yet the paragraph-to-paragraph progression often reflects dimension switching rather than inferential advancement. 
In Case~2 (resigning to pursue a fully funded Princeton PhD), the response first develops a risk-oriented block. It discusses financial pressure (cost of living, travel, and loss of income) and psychological adaptation (uncertainty, stress, and cultural adjustment). 
This setup introduces a set of constraints that would need to be reconciled for the argument to progress.

However, the response does not derive such a criterion. It does not summarize the risks into a decision rule (e.g., whether expected long-term gains outweigh short-term costs given one’s financial buffer and risk tolerance). Instead, it abruptly pivots to a benefit-oriented block about long-term career ceilings and academic networks.

The transition is therefore only superficial. Connective phrases like “from a career development perspective” and “therefore, in the long run” provide rhetorical flow, but they do not establish a discourse relation that links the two blocks. In particular, the text never explains how the earlier constraints enter the later argument (as assumptions, thresholds, or explicit trade-off terms), nor does it clarify which factors dominate under which scenario. As a result, the response reads like a coherent checklist of considerations rather than a step-by-step decision argument with explicit dependencies and a resolved trade-off.

\noindent\textbf{(3) Unsupported Warrants.}
Analytical claims require explicit warrants because, without an inferential bridge, readers cannot verify whether the conclusion follows from the presented evidence, and fluent prose can mask unsupported causal leaps.
In Case~3 (the ``Cell Lysis Device Market'' report), the model repeatedly treats high-level drivers (e.g., ``biomedical industry expansion,'' ``policy support,'' ``precision medicine,'' and ``technology progress'') as sufficient evidence for the conclusion that ``market demand increases.'' However, the report rarely specifies the intermediate links that would make these causal statements verifiable. For example, the claim that ``biomedical industry expansion has pushed market demand'' leaves the mechanism under-specified: it does not identify which downstream workflows are expanding (e.g., nucleic-acid testing, protein extraction, or drug discovery), why cell lysis becomes a bottleneck in those workflows (e.g., throughput or reproducibility constraints in sample preparation), or through which purchasing channel demand materializes (e.g., capacity expansion, automation upgrades, or replacement cycles). As a result, the statement functions as a correlation-like association rather than a justified causal pathway.

The same failure is evident for ``policy support.'' The report asserts that policy ``creates a favorable environment'' and ``promotes expansion,'' yet it does not articulate any concrete instrument that could plausibly translate into device procurement (e.g., research grants that expand equipment budgets, platform-building programs that trigger centralized purchasing, or standardization/regulatory changes that incentivize automated sample-processing pipelines). Instead, the text uses policy as rhetorical cause and jumps directly to market growth, leaving the reader unable to check whether the conclusion follows. 
Overall, the report matches the tone of professional analysis---enumerating drivers with confident language---but omits the inferential bridge that makes the argument auditable, and therefore reads as a list of plausible factors rather than a supported causal account.

\noindent\textbf{Implication.}
Across cases, the most damaging failures are not surface fluency errors but missing global constraint reconciliation, under-specified discourse relations between adjacent sections, and absent warrants for key inferences. 
These observations motivate evaluation and training signals that explicitly target consistency, warranted inference, and discourse-level linkage in long-form analytical reports, which are precisely the failure modes emphasized by the ReportLogic taxonomy.

\onecolumn
{
\small

\begin{tcolorbox}[
  enhanced, breakable,
  colback=softGray,
  colframe=badRed!60,
  title=\dimtag{Case 1: Internal Consistency (Global constraints violated)},
  fonttitle=\bfseries,
  boxrule=0.6pt, arc=2pt, left=6pt, right=6pt, top=6pt, bottom=6pt]
\textbf{Query (abbr.).} Collect and summarize the real income and financial status of China’s nine social strata; analyze the defining traits, size, and financial capacity of the middle class.

\vspace{4pt}
\textbf{Trigger span (excerpt; translated from original text).}
Stratum 2: Upper-middle-class households account for \badhl{1\%} of all households (about 3.2 million households), with net assets around 12.12 million RMB and annual net income around 570k RMB.\\
Stratum 3: Ordinary middle-class households account for \badhl{10\%} of all households (about 32 million households), with net assets around 3.42 million RMB and annual net income 189k RMB.\\
Stratum 4: ``Xiaokang'' households account for \badhl{45\%} of all households (about 144 million households), with net assets around 600k RMB and annual net income 96k RMB.\\
Stratum 5: Subsistence households are described as \badhl{``the lower half of 75\% of all households''} (about 80 million households), with net assets 117k RMB and annual net income 39k RMB.\\
Strata 6--9: Basic survival groups together account for about \badhl{25\%} of all households (annual income typically below 20k RMB).\\
(The report also assigns non-zero shares to Stratum 1 and other groups.)

\vspace{4pt}
\textbf{Failure diagnosis (what goes wrong).}
The problem is not whether any single number is individually plausible; it is a violation of \emph{global accounting constraints}.
The excerpt mixes (i) shares that are framed as fractions of \emph{all households} (e.g., \badhl{45\%}, \badhl{10\%}, \badhl{1\%}, plus \badhl{25\%}) with (ii) an additional statement whose \emph{denominator and nesting are unclear} (the \badhl{``lower half of 75\%''} phrasing).
Because the report never specifies whether strata are mutually exclusive, nested, or overlapping—and never performs any reconciliation—the strata cannot be verified as a coherent partition.

\vspace{4pt}
\textbf{Missing bridge / Minimal fix.}
\fixhl{Explicitly define whether the nine strata are mutually exclusive; standardize the denominator (all households vs.\ a subset); and provide a cumulative sum-check to ensure totals equal 100\% (up to rounding).}
\end{tcolorbox}

\vspace{8pt}

\begin{tcolorbox}[
  enhanced, breakable,
  colback=softGray,
  colframe=badRed!60,
  title=\dimtag{Case 2: Local Coherence (``smooth but checklist-like'' progression)},
  fonttitle=\bfseries,
  boxrule=0.6pt, arc=2pt, left=6pt, right=6pt, top=6pt, bottom=6pt]
\textbf{Query (abbr.).} How should one think about resigning to pursue a fully funded Princeton PhD offer?

\vspace{4pt}
\textbf{Trigger span (excerpt; translated from original text; adjacent paragraphs).}
\emph{However, resigning to study abroad is not without risks. First, financial factors cannot be ignored. Even with full funding, living costs, international travel expenses, and other potential costs still require careful planning... In addition, resigning means giving up the current income source, which may create short-term financial pressure. Therefore, before making this decision, one must comprehensively assess one’s financial situation and create a detailed budget plan to ensure basic living needs during the study period.}\\[2pt]
\emph{\badhl{From a career development perspective}, whether resigning to study abroad is worthwhile depends on personal career goals and the characteristics of the industry... Princeton’s PhD program not only provides high-quality academic training, but also helps students build extensive academic networks, which is important for future career development... Moreover, Princeton’s reputation and influence can offer more job opportunities and higher starting salaries.}

\vspace{4pt}
\textbf{Failure diagnosis (why coherence breaks locally).}
The paragraph boundary marks a \emph{dimension switch} (short-term risks $\rightarrow$ long-term upsides) rather than a stepwise progression.
Although the second paragraph begins with \badhl{“From a career development perspective”}, it does not state how the prior risk analysis enters the subsequent argument (as assumptions, thresholds, or explicit trade-off terms).
Crucially, the text never introduces an intermediate decision criterion—e.g., “if the financial runway is at least $X$ months and the stress tolerance is above $Y$, then the short-term risks are acceptable.”
Without such a bridge, the argument does not accumulate: each paragraph contributes a relevant factor, but the reader is not shown how the previous paragraph becomes a premise for the next.
The result is “smooth but checklist-like” writing: cohesive at the surface, but locally under-connected.

\vspace{4pt}
\textbf{Missing bridge / Minimal fix.}
\fixhl{Add one bridging sentence that summarizes the risks into an operational condition (financial buffer / risk tolerance threshold), and open the next paragraph by stating that long-term benefits dominate only when that condition holds.}
\end{tcolorbox}

\vspace{8pt}

\begin{tcolorbox}[
  enhanced, breakable,
  colback=softGray,
  colframe=badRed!60,
  title=\dimtag{Case 3: Warrants \& Causal Reasoning (Unsupported causal drivers)},
  fonttitle=\bfseries,
  boxrule=0.6pt, arc=2pt, left=6pt, right=6pt, top=6pt, bottom=6pt]
\textbf{Query (abbr.).} Report on the global and China market status and future trends of cell lysis devices.

\vspace{4pt}
\textbf{Trigger span (excerpt; translated from original text).}
\emph{The growth of the global cell lysis device market is mainly driven by the following factors. First, continued progress in life-science research—especially rapid development in genomics, proteomics, and cell biology—raises higher requirements for efficient and reliable lysis equipment. Second, \badhl{the expansion of the biomedical industry}, especially increasing demand in biopharmaceuticals and molecular diagnostics, \badhl{has driven market demand for cell lysis devices}. In addition, \badhl{technological progress is also an important driver of market growth}. In recent years, emerging technologies such as automation, microfluidics, and nanotechnology applied to cell lysis devices not only improve device performance but also expand their application scope.}

\vspace{4pt}
\textbf{Failure diagnosis (unsupported warrants).}
Warrants matter because they make causal claims \emph{checkable}: they specify \emph{why} a purported driver should increase demand, rather than merely asserting correlation.
In this excerpt, causal language is used repeatedly (\badhl{“industry expansion” $\rightarrow$ “driven demand”}, \badhl{“technology progress” $\rightarrow$ “market growth”}) but the mechanism is left implicit.
The text does not explain which downstream workflows scale up (e.g., diagnostics sample preparation, protein extraction in biopharma), why lysis becomes a throughput bottleneck, or why that bottleneck translates into additional device procurement.
Likewise, “technological progress” is stated as a driver without clarifying whether it increases throughput, reduces unit cost, enables new use cases, or changes adoption incentives.
As a result, the report matches the tone of professional market analysis but omits the inferential steps needed to verify the causal chain.

\vspace{4pt}
\textbf{Missing bridge / Minimal fix.}
\fixhl{Provide at least one complete mechanism chain (industry expansion $\rightarrow$ higher sample volume/throughput $\rightarrow$ more lysis operations $\rightarrow$ capacity bottleneck $\rightarrow$ increased procurement), and make explicit how “technology progress” affects demand (e.g., cost reduction / throughput gain / new use cases / regulatory or workflow constraints).}
\end{tcolorbox}

\captionof{figure}{Bad-case visualization for Deep Research-style reports.
\badhl{Red highlights} mark trigger spans where a rubric constraint is violated;
\fixhl{blue highlights} indicate the missing logical element required by that dimension to make the report auditable.}
\label{tab:bad_cases}
}
\twocolumn

\begin{figure}[t]  
\setlength{\abovecaptionskip}{0cm}
\setlength{\belowcaptionskip}{0cm}
    \centering    
    \includegraphics[width=0.95\linewidth]{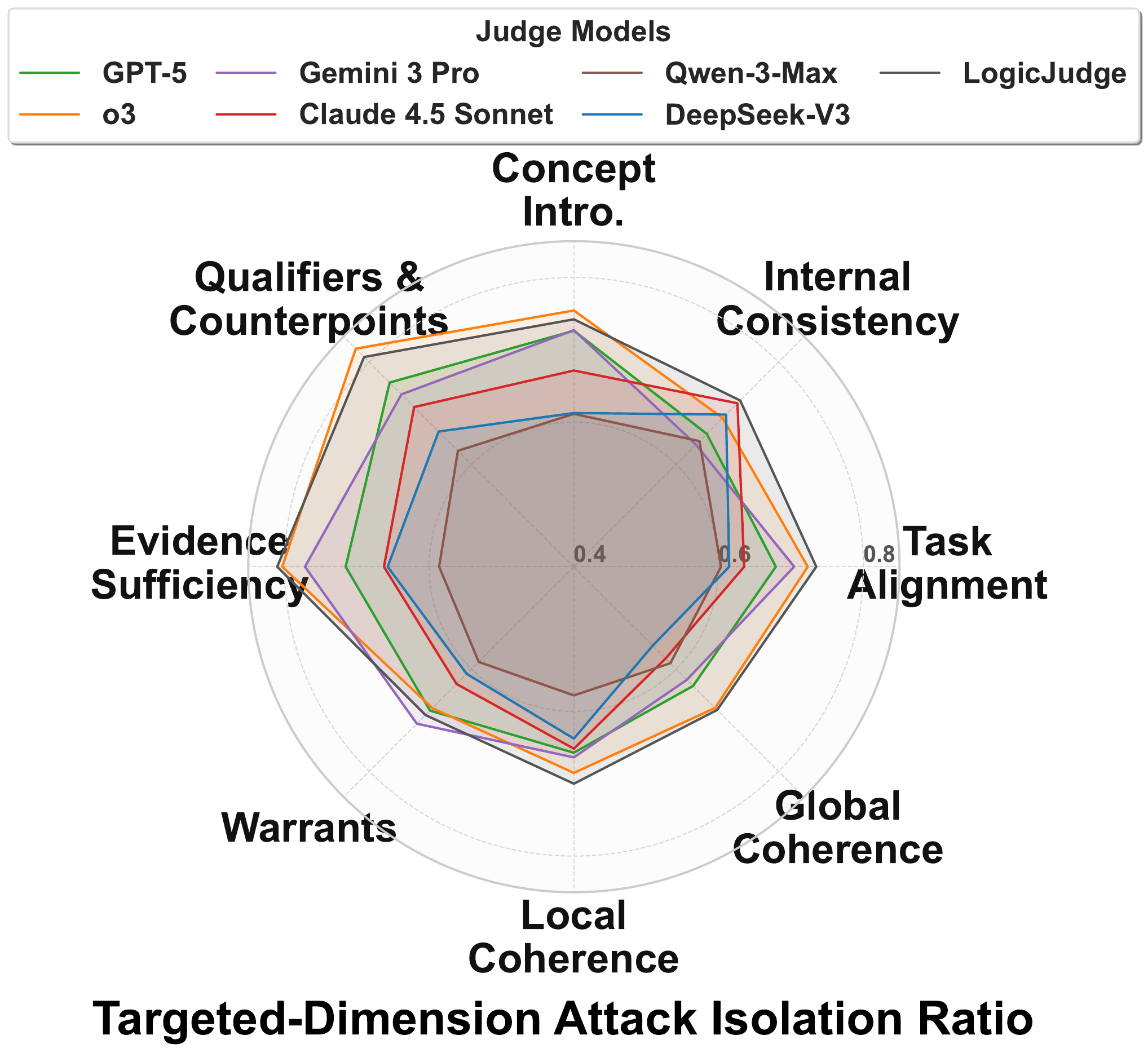}
    \caption{Isolation Ratio (IR) under targeted-dimension attacks.}
    \label{fig:aspect_attack_ir}
\end{figure}

\subsection{Attack Analysis Details}
\label{app:attack}
This appendix provides additional details for the attack-based evaluation in RQ3, including adversarial construction protocols, metrics, and extended interpretations.
The main paper emphasizes Attack Success Rate (ASR) as the primary robustness indicator; here we additionally report Isolation Rate (IR) for the Targeted-Dimension Attack to analyze whether a judge can localize penalties to the attacked rubric dimension.

\paragraph{Adversarial Construction.}
We randomly sample $N=300$ ReportLogic instances, each consisting of a Deep Research query and an associated model response $x$.
For each instance, we generate an adversarial candidate $x_{\mathrm{adv}}$ under one of two attack suites using Gemini-2.5-Pro (selected as a high-performing model on our logical-quality leaderboard).
We then perform manual screening with two trained annotators to ensure each candidate satisfies the suite-specific constraints (targeted degradation for the Targeted-Dimension Attack, or logical equivalence with only surface manipulation for the Bias-Type Attack), resolving disagreements by discussion.
The full prompt templates used for adversarial generation are released in our anonymous repository.
After finalizing $x_{\mathrm{adv}}$, we apply the same context-aware rubric generation pipeline as in Section~\ref{sec:bench_data} to produce instance-specific inspection items for the paired comparison, and then evaluate different judges by asking them to compare $x$ against $x_{\mathrm{adv}}$ under these rubrics.

The details of two kinds of attack are as follows:
\begin{itemize}[leftmargin=*]
    \item \textbf{Targeted-Dimension Attack (Sensitivity Probe).}
    This suite tests whether a judge can penalize a \emph{localized} logical defect within a designated rubric dimension.
    We construct $x_{\mathrm{adv}}$ by injecting an error that targets only the chosen dimension (e.g., removing a warrant in a causal chain or introducing a local contradiction), while keeping other aspects unchanged by construction.
    A reliable judge should consistently prefer the original response $x$ over the attacked version $x_{\mathrm{adv}}$.

    \item \textbf{Bias-Type Attack (Robustness Probe).}
    This suite tests whether a judge is influenced by superficial presentation cues when the adversarial response is \emph{logically equivalent} to the original.
    We instantiate five bias types:
    (1) \textit{Length Bias}: inflating verbosity via redundant paraphrasing without adding new information;
    (2) \textit{Structure Bias}: adding outline-style scaffolding (e.g., light headers, numbering, discourse signposts, and brief transitions or summaries) while preserving the original claims and order;
    (3) \textit{Qualifier Wording}: strengthening cautious hedges and generic limitation statements without introducing new conditions or counterexamples;
    (4) \textit{Evidence Illusion}: labeling and listing existing statements as ``evidence'' (optionally with light aggregation phrasing) without introducing new supporting content; and
    (5) \textit{Causal Display}: mechanically making an existing causal chain explicit (e.g., rewriting implicit links into ``because--therefore'' form and step-by-step presentation) without adding new intermediate claims, mechanisms, or evidence.

\end{itemize}

\paragraph{Screening Protocol.}
Raw adversarial candidates generated by Gemini-2.5-Pro are manually filtered by two trained annotators before being used in the evaluation.
Each candidate $x_{\mathrm{adv}}$ is independently assessed against three criteria, and is retained only if both annotators agree it satisfies all three:
(1) \textbf{Dimension Isolation.}
    For Targeted-Dimension attacks, the injected defect must affect only the designated rubric dimension, without introducing collateral damage to other dimensions (e.g., a \textit{Warrants} attack must not simultaneously alter task alignment or internal consistency).
    For Bias-Type attacks, the manipulation must be confined to surface cues (length, structure, wording) without changing the underlying claims, evidence, or reasoning chain on any dimension.
(2) \textbf{Minimal Edits.}
    The adversarial version must preserve the original response as much as possible outside the attack target.
    For Targeted-Dimension attacks, annotators verify that unaffected content remains lexically and semantically close to the original.
    For Bias-Type attacks, annotators verify that no new propositions, evidence, or qualifications are introduced; only surface re-packaging is allowed.
(3) \textbf{Bounded Strength.}
    The defect must be perceptible but not catastrophic: a Targeted-Dimension attack should degrade the targeted aspect to a clearly sub-standard level without producing an obviously incoherent report, and a Bias-Type attack should apply a realistically plausible surface manipulation rather than an exaggerated distortion.
    This bound ensures that judge vulnerabilities reflect subtle reasoning shortcuts rather than trivial edge cases.
Candidates failing any criterion are regenerated (for Targeted-Dimension attacks where the defect is too broad, too weak, or leaks into other dimensions) or discarded (for Bias-Type attacks that introduce new content).
Disagreements between the two annotators are resolved via discussion until consensus is reached.

\paragraph{Representative Examples.}
To make the two attack families concrete, Table~\ref{tab:attack_examples} presents five paired excerpts showing reports before and after attack, drawn from our actual attack set. The first three illustrate \emph{Targeted-Dimension} attacks that weaken a specific logical dimension (Qualifiers, Internal Consistency, Concept Introduction), while the last two illustrate \emph{Bias-Type} attacks that preserve the underlying argument but inject a surface signal (report-style numbering, explicit causal connectives). In each row, modifications relative to the original are highlighted in bold, omitted passages are marked with ``\ldots''.

\begin{table*}[t]
\centering
\small
\renewcommand{\arraystretch}{1.25}
\setlength{\tabcolsep}{4pt}
\caption{Representative excerpts showing reports before and after attack. In each row, bold marks the spans affected by the attack---in the original column, the passages that are modified; in the attacked column, the resulting modifications. ``\ldots'' denotes omitted context. All five examples are drawn from real attack outputs in our benchmark.}
\label{tab:attack_examples}
\begin{tabular}{p{2.2cm}|p{6.4cm}|p{6.4cm}}
\toprule
\textbf{Attack} & \textbf{Original (excerpt)} & \textbf{Attacked (excerpt)} \\
\midrule

\multicolumn{3}{l}{\textit{Targeted-Dimension attacks}} \\
\midrule

Qualifiers \newline \& Counterpoints &
\ldots The long-term implication of DLT is the \textbf{potential} creation of a new financial market infrastructure that is more transparent, efficient, and interoperable, \textbf{though significant challenges related to scalability, regulation, and energy consumption remain to be fully addressed}. \ldots AI adoption\ldots \textbf{raises concerns regarding job displacement and ethical considerations}; \textbf{the opacity of ``black box'' algorithms poses challenges for transparency and accountability}\ldots &
\ldots The long-term implication of DLT is the \textbf{inevitable} creation of a new financial market infrastructure that is more transparent, efficient, and interoperable, \textbf{with only minor adjustments likely to be needed over time}. \ldots AI adoption\ldots \textbf{will deliver substantial benefits across the industry}; \textbf{modern algorithms consistently provide reliable outcomes in lending and insurance}\ldots \\

\midrule

Internal \newline Consistency  &
Open Banking is a regulatory and technological framework that mandates financial institutions to securely share customer data with third-party providers (TPPs) through standardized APIs. \ldots This fosters unprecedented competition and innovation\ldots &
Open Banking is a regulatory and technological framework\ldots through standardized APIs. \textbf{In its narrow sense, it is primarily about enabling data portability and account access for licensed payment and information service providers.} \ldots \textbf{In practice, this has increasingly meant that Open Banking is treated as a broad umbrella for almost any form of API-based financial integration, including services that go well beyond data access and into areas like full-service embedded lending and insurance.} \\

\midrule

Concept \newline Introduction \newline \& Logical  \newline Transition  &
The global financial industry is presently undergoing a transformation of unprecedented scale and velocity, \textbf{driven by a confluence of technological advancement, evolving regulatory frameworks, and shifting consumer expectations}\ldots \textbf{The catalysts\ldots include the lingering structural lessons from the 2008 global financial crisis, the rapid proliferation of digital connectivity, and\ldots new consumers\ldots} \ldots leveraging modern API protocols \textbf{to facilitate seamless communication between disparate banking institutions}\ldots &
The global financial industry is presently undergoing a transformation of unprecedented scale and velocity, \textbf{with RegTech, tokenization, and embedded finance now redefining the sector}. \ldots \textbf{[background on 2008 crisis and digital connectivity moved to a later paragraph]}\ldots leveraging modern API protocols \textbf{for inter-bank messaging}\ldots \\

\midrule
\multicolumn{3}{l}{\textit{Bias-Type attacks}} \\
\midrule

Structure \newline Bias &
\textbf{Chess Progression and Technical Analysis} \newline In this game, Ke Jie played white\ldots captured the half-point victory. From a technical standpoint, the reasons for Ke Jie's win can be grouped into three points: accurate judgment of complex positions, decisive use of sacrifice tactics, and flawless endgame execution. &
\textbf{II. Chess Progression and Technical Analysis: A Complete Arc from Opening through Middlegame to Endgame} \newline In this game, Ke Jie played white\ldots captured the half-point victory. From a technical standpoint, the reasons for Ke Jie's win can be grouped into three points: accurate judgment of complex positions, decisive use of sacrifice tactics, and flawless endgame execution, \textbf{laying a clear technical foundation for the subsequent discussion of strategy and psychology}. \\

\midrule

Causal \newline Display \newline Bias &
The global financial industry is presently undergoing a transformation of unprecedented scale and velocity, driven by a confluence of technological advancement, evolving regulatory frameworks, and shifting consumer expectations. This period of intense innovation represents a fundamental departure from the incremental changes that characterized the late 20th century\ldots &
\ldots driven by a confluence of technological advancement, evolving regulatory frameworks, and shifting consumer expectations. \textbf{Because these forces are acting together rather than in isolation,} this period of intense innovation represents a fundamental departure from the incremental changes that characterized the late 20th century\ldots \\

\bottomrule
\end{tabular}
\end{table*}

\paragraph{Evaluation Metrics.}
To quantify model performance, we employ two key metrics:
(1) \textbf{Attack Success Rate (ASR):} 
    This measures the judge's vulnerability to adversarial manipulation. 
    For Targeted-Dimension attacks, ASR is the proportion of cases where the judge erroneously prefers the degraded $x_{\mathrm{adv}}$ or issues a tie on the targeted dimension (expecting a clear penalty on $x_{\mathrm{adv}}$). A tie is counted as a success here because the attack objective is to prevent the judge from clearly penalizing the injected defect, and a tie already means the judge fails to prefer the clean report on the attacked aspect.
    For Bias-Type attacks, ASR denotes the frequency with which the judge favors the manipulated text (e.g., the longer one) over the logical original (expecting resistance). A tie is counted as non-success in this setting, because the attack objective is to actively flip the judge toward the attacked side. A tie indicates that the judge does not prefer the attacked output and is therefore not deceived by the surface manipulation.
(2) \textbf{Isolation Rate (IR):} 
    Exclusively calculated for the Targeted-Dimension Attack, this metric assesses the selectivity of the judge's critique.
    This metric verifies whether the penalty imposed by the judge is confined to the attacked dimension. 
    A high IR indicates that the judge accurately localizes the error (e.g., penalizing only \textit{Warrants \& Causal Reasoning} for a broken chain) without reducing scores for unaffected dimensions (e.g., \textit{Task Alignment}), thereby demonstrating precise, disentangled reasoning capabilities rather than a generalized negative halo effect.
    
\paragraph{Extended Interpretation.}
Beyond the main observations in Section~\ref{subsec:exp_attack}, the targeted-dimension attack further reveals a clear separation between error localization and attack robustness.
Figure~\ref{fig:aspect_attack_ir} shows that o3 achieves the highest IR across most dimensions, indicating that when a localized defect is introduced, its penalties are largely confined to the attacked aspect rather than spilling over broadly.
This suggests that o3 more consistently follows a dimension-conditioned evaluation procedure, producing sharper aspect attribution.

However, high isolation does not necessarily imply low attack success.
As shown in Figure~\ref{fig:target_dim_attack}(a), o3 still exhibits relatively elevated ASR on several dimensions compared to GPT-5 and Gemini~3~Pro.
Taken together, these results suggest that o3 can localize the defect once it affects its judgment, but may still be easier to flip under certain targeted degradations.
A plausible explanation is that reasoning-oriented judges may attempt to preserve a coherent interpretation by implicitly reconstructing missing bridges.
This tendency can reduce sensitivity to attacks that weaken explicit support relations, making the model more likely to accept a subtly degraded answer, while still assigning the resulting penalty primarily to the most relevant dimension.

In contrast, Qwen-3-Max exhibits the lowest IR overall, and DeepSeek-V3 also shows consistently low IR, indicating substantial cross-dimension spillover.
This pattern is consistent with impression-driven scoring: once a response is judged worse, multiple aspect ratings are jointly reduced to match the global preference, producing a halo effect rather than precise attribution.
Overall, IR provides a complementary diagnostic signal to ASR, distinguishing judges that are vulnerable because they are easily flipped from those that are vulnerable because they cannot precisely localize the source of degradation.

\paragraph{LogicJudge: Extended Analysis.}
We additionally examine LogicJudge as an evaluation-aligned baseline judge trained on ReportLogic supervision.
Overall, LogicJudge exhibits relatively low ASR across most targeted dimensions, and achieves strong error localization: in Figure~\ref{fig:aspect_attack_ir}, it ranks among the top judges in IR, particularly on \textit{Warrants \& Causal Reasoning} and \textit{Concept Intro}, where it is consistently second only to the most localized judge.
This suggests that once a defect is recognized, LogicJudge tends to attribute the penalty to the intended rubric aspect rather than triggering broad cross-dimension spillover.
Nevertheless, two failure modes remain salient.
First, LogicJudge is more affected by \textit{Length Bias}.
A plausible explanation is that our training objective is pairwise and rubric-conditioned: when two responses are logically equivalent, the decision boundary becomes narrow and the model must rely on limited comparative signals.
Longer rewrites systematically increase redundancy and restatement, which can reduce apparent ambiguity and make support relations appear easier to trace, thereby shifting borderline comparisons even without changing the underlying claims.
Second, LogicJudge shows higher ASR under \textit{Local Coherence} attacks.
Local coherence defects rarely manifest as a single clearly wrong sentence; instead, they arise from subtle cross-sentence mismatches (e.g., drifting conditions, inconsistent references, or quietly altered claim scopes) that only become evident when aligning information across multiple sentences.
In our distillation setup, the supervision includes dimension-level pairwise judgments with brief, high-level rationales, but it may not provide fine-grained, sentence-level attributions that explicitly pinpoint the exact spans responsible for the inconsistency.
As a result, when coherence defects are implicit and distributed, they may be insufficient to trigger a decisive preference flip, making it harder for the model to learn a stable boundary for detecting such fine-grained inconsistencies.

\section{Artifact Documentation and Ethical Considerations} \label{app:ethics_artifacts} \paragraph{Artifacts and attribution.} This work relies on a combination of publicly available datasets, model-generated reports, and evaluation artifacts introduced in this paper. All external datasets and models used in our experiments are properly cited in the main paper. We do not claim ownership over any third-party artifacts, and we follow the original terms under which these resources are released. \paragraph{Licensing and terms of use.} All datasets used in this study are either publicly accessible for research purposes or consist of model-generated content. Our use of these resources is strictly limited to non-commercial research and evaluation. We do not redistribute raw data that is subject to restrictive licenses; instead, we report aggregate statistics, evaluation outcomes, and derived annotations consistent with common academic practice. \paragraph{Intended use and consistency.} The artifacts created in this work---including the evaluation benchmark ReportLogic and judge model LogicJudge---are intended solely for research and diagnostic evaluation of long-form logical quality. Our use of existing datasets and model outputs is consistent with their intended research use. Derived annotations and judgments are not designed for deployment in downstream decision-making systems or real-world moderation settings. \paragraph{Personally identifiable information and offensive content.} The data analyzed in this study consist of public or model-generated text and may contain natural language references typical of open-domain content. We do not collect, analyze, or model personally identifiable information. User identifiers and metadata are removed or anonymized where applicable, and the evaluation focuses exclusively on report-level logical structure rather than individual attributes or identities. 
Potentially offensive content is not targeted or amplified. During dataset curation and human annotation, we filter and remove instances containing offensive content; remaining data are used solely for research evaluation under standard anonymization procedures.

\section{Human Subjects Including Annotators}
\label{app:human}

This study involves human annotators for constructing and validating the ReportLogic benchmark. 
All annotation procedures were designed to follow established ethical standards for research involving human subjects and to pose minimal risk to participants.

\paragraph{Instructions to Annotators.}
All annotators received comprehensive written guidelines prior to participation. For each annotation task, we first provided a plain-language overview of the study goal and the specific judgment to be made, so that annotators could build an accurate mental model of what constitutes a valid comparison. We then supplied precise definitions of the evaluation criteria and the decision rules for each dimension, together with multiple concrete examples illustrating both acceptable and unacceptable judgments for each possible outcome. The instructions also specified quality-control expectations (e.g., careful reading of both reports, avoiding reliance on superficial cues, and following the prescribed procedure before submitting a verdict). Annotators were informed that their labels would be used for research purposes. No deception was involved, and the task consisted solely of expert assessment of model-generated text.

\paragraph{Annotation Workflow and Quality Assurance.}
Each instance was independently annotated by three annotators following the same rubric-guided procedure. Annotators first reviewed the query and the provided context, then compared the two reports dimension by dimension, and finally selected a three-way verdict for each dimension and overall preference. We monitored inter-annotator agreement throughout the annotation process and used it as a primary quality signal.

For instances where the three annotations did not yield a majority decision (i.e., three-way disagreement) or where the disagreement indicated substantive ambiguity rather than minor differences in interpretation, we applied an adjudication step. Specifically, an expert adjudicator (a domain expert Ph.D. researcher with prior experience in logical evaluation and analytic writing) re-examined the full instance, including the rubric, both reports, and the annotators' rationales. The adjudicator either (i) selected the final verdict when one option was clearly better supported under the rubric, or (ii) initiated a short reconciliation procedure by documenting the decisive rubric criteria and resolving the disputed points to ensure the final label was consistent with the written guidelines. This process produced a single consolidated gold annotation for each instance.

\paragraph{Recruitment and Compensation.}
Annotators were recruited by the research team based on prior experience with analytical writing and logical evaluation tasks.
In total, 25 annotators participated in the study.
All annotators hold at least a master’s degree in relevant fields, including linguistics, literature, philosophy, or technical writing.
Annotators were compensated on a per-task basis at a rate equivalent to approximately \$28 USD per annotated instance, reflecting the substantial cognitive effort required for fine-grained, rubric-guided logical assessment of long-form reports.
Compensation was determined in advance and was independent of annotation outcomes.

\paragraph{Consent and Data Use.}
All annotators provided informed consent prior to participation.
The instructions explicitly described how the annotated data would be used, stored, and reported.
Annotations were analyzed only in aggregate form, and no personally identifiable information was collected, stored, or released.
Annotators were informed that participation was voluntary and that they could withdraw at any time without penalty.

\paragraph{Ethics Review.}
The annotation protocol and data collection procedures were reviewed and approved through the internal ethics and compliance review process of the authors' research team.
The study involves minimal risk, as it consists solely of expert evaluation of text generated by language models.

\paragraph{Annotator Characteristics.}
Annotators are adult participants (18+), fluent in the language of the annotated content, and selected based on demonstrated expertise rather than demographic attributes.
Basic professional characteristics (educational background and domain expertise) were considered for recruitment, while no sensitive personal attributes were collected or used.

\section{Use of AI assistants}
AI assistants were used exclusively for language polishing, such as grammar correction, clarity improvement, and minor stylistic edits.
They were not used for idea generation, methodological design, experimental execution, result analysis, or interpretation.
All scientific contributions and conclusions remain the responsibility of the authors.

\section{Prompt Template}
\label{app:prompts}
\onecolumn
{
\small
\begin{tcolorbox}[
  enhanced, breakable,
  colback=softGray,
  colframe=badRed!60,
  title={Query Filtering Prompt (Deep Research Suitability)},
  fonttitle=\bfseries,
  boxrule=0.6pt, arc=2pt, left=6pt, right=6pt, top=6pt, bottom=6pt]

\textbf{Prompt:}
You are a rigorous content suitability judge for a Retrieval-Augmented Generation (RAG) based Deep Research system.
Given a query, decide whether it is suitable for Deep Research style long-form reporting.
Always output exactly one JSON dictionary on a single line, and output nothing else.

\vspace{6pt}
\textbf{Overall Objective.}
ACCEPT queries that typically (i) require consulting up-to-date or specialized external resources or systematically synthesizing substantial background knowledge, theories, and cases,
(ii) support analysis across multiple dimensions rather than a single fact or definition,
and (iii) are best answered as a structured long-form report, usually extendable to 1500--2000 words or more with multiple sections.
Queries better suited to short answers, encyclopedic explanations, simple listings, or common-sense replies should generally be REJECTED.

\vspace{8pt}
\textbf{RAG / Deep Research Suitability Criteria.}
Output decision ``ACCEPT'' only if the query clearly satisfies multiple criteria below (e.g., at least three).
If it only weakly satisfies one or two, output ``REJECT''.

\vspace{4pt}
\begin{enumerate}\setlength{\itemsep}{4pt}\setlength{\leftskip}{0pt}
  \item \textbf{Strong retrieval dependence:}
  A reasonable answer requires consulting external sources such as academic papers, industry reports, empirical data, technical documentation, news, regulations, standards, or real-world cases, or systematically organizing such information.

  \item \textbf{Multi-source synthesis:}
  Answering requires integrating information from multiple sources or perspectives rather than citing a single fact or data point, and often involves comparing viewpoints, methods, metrics, or solutions.

  \item \textbf{Multi-dimensional decomposability:}
  The query can be naturally decomposed into multiple analytical dimensions (e.g., background, core concepts, current approaches, trade-offs, cases or data comparisons, risks, and future trends).
  If it essentially asks to ``look up one thing'', output ``REJECT''.

  \item \textbf{Complexity and openness:}
  The query is not a simple yes/no question or a mechanical listing task.
  It requires reasoning, trade-off analysis, or argumentation, often involving ``why'', ``with what impact'', or ``how to design, optimize, or balance''.

  \item \textbf{Long-form report suitability:}
  A reasonable answer would naturally include multiple sections (e.g., motivation, foundations, existing approaches, case studies or comparisons, challenges, and recommendations).
  If it can be fully addressed in one or two short paragraphs or a short list of steps, output ``REJECT''.
\end{enumerate}

\vspace{8pt}
\textbf{Hard Constraints (Must Reject).}
If any condition holds, output decision ``REJECT'' regardless of the criteria above:

\vspace{4pt}
\begin{itemize}\setlength{\itemsep}{4pt}
  \item Sexual content, extreme violence, self-harm, suicide, or illegal activities.
  \item Real-world politics, governments, political parties, elections, diplomacy, or ideology, regardless of stance or tone.
  \item Attacks, denigrates, or stigmatizes a specific country, region, or city, explicitly or implicitly, including suggestive or veiled framing.
  \item Closed-form tasks solvable without external retrieval (e.g., pure mathematical proofs, basic programming exercises, or logic puzzles).
  \item Casual chat, personal life issues, emotional venting, or family or relationship disputes.
  \item Purely definitional or short FAQ-style queries that do not support long-form structured analysis.
  \item Queries that are inherently short and cannot be reasonably expanded into a multi-section analytical report.
\end{itemize}

\vspace{8pt}
\textbf{Output Format (JSON dictionary, two keys).}
Return exactly one JSON dictionary on a single line, with exactly two keys: ``decision'' and ``reason''.
The value of ``decision'' must be either ``ACCEPT'' or ``REJECT''.
The value of ``reason'' must be a brief justification. For ``REJECT'', state the primary reason (e.g., too simple, definitional, political, or not retrieval-dependent).

\vspace{6pt}
\textbf{Input.}
Query: \{query\}

\vspace{6pt}
\textbf{Example output.}
\{``decision'':``ACCEPT'',``reason'':``Requires external retrieval and multi-source synthesis; supports multi-dimensional analysis and naturally fits a long, structured report.''\}

\end{tcolorbox}

\captionof{figure}{Prompts used to filter open-domain queries for Deep Research suitability.}
\label{fig:query_filter}
}
\twocolumn

\onecolumn
{
\small
\begin{tcolorbox}[
  enhanced, breakable,
  colback=softGray,
  colframe=badRed!60,
  title={Report Generation Prompt for Deep Research},
  fonttitle=\bfseries,
  boxrule=0.6pt, arc=2pt, left=6pt, right=6pt, top=6pt, bottom=6pt]

\textbf{Prompt:}
You are a domain expert writing a formal analytical REPORT based on your own expertise and the provided relevant documents retrieved via a Retrieval-Augmented Generation (RAG) pipeline.

\vspace{6pt}
\textbf{Objective.}
Write a comprehensive and detailed report, not a summary or abstract.
Your task is to reason deeply, synthesize evidence, and produce a coherent long-form narrative that could be published as an expert analytical report or white paper.

\vspace{8pt}
\textbf{Writing Requirements (STRICT).}
Paragraphized prose only. Each paragraph must contain at least four complete sentences.
Absolutely no bullet points, numbered lists, outlines, or tables.
Maintain a formal, academic, and explanatory tone throughout.
Logical flow is required: every section must connect naturally to the previous one with explicit transitions.
Prioritize depth over breadth. For every major point, explain mechanisms (why and how), contextual background, and implications.
Length requirement: for English or non-CJK queries, produce at least 3{,}000 words. For Chinese (CJK) queries, produce at least 6{,}000 characters excluding spaces. If uncertain, follow the CJK requirement.

\vspace{8pt}
\textbf{Expansion Policy (When RAG Evidence Is Insufficient).}
If the retrieved RAG documents lack coverage, contain gaps, or address the topic only partially, expand the discussion using your domain knowledge while ensuring that all additional content remains reasonable, verifiable, and logically consistent with the RAG context.
Never fabricate facts, numbers, or citations.

\vspace{8pt}
\textbf{Input.}
Query: \{new\_query\}

RAG Documents (with IDs): \{docs\_block\}

\vspace{6pt}
\textbf{Output.}
Produce a single, continuous REPORT that strictly follows all requirements above.
Begin the report immediately below this instruction without any prefatory remarks.

\end{tcolorbox}

\captionof{figure}{Prompt used for Deep Research style long-form report generation.}
\label{fig:gen_prompts}
}
\twocolumn

\onecolumn
{
\small
\begin{tcolorbox}[
  breakable,
  colback=softGray,
  colframe=badRed!60,
  title={Context-aware Rubric Generation Prompt},
  fonttitle=\bfseries,
  boxrule=0.6pt,
  left=6pt, right=6pt, top=6pt, bottom=6pt
]

\textbf{Prompt:}
You are a logic analysis expert.
Your task is: given a QUERY and two response texts (A and B), generate a logic evaluation rubric for human reviewers to compare A and B across eight logical aspects.

\medskip
\textbf{Three-layer Logic Framework.}
This evaluation is grounded in classical logic and argumentation analysis, and decomposes text-level logic into three layers:
(1) \textbf{Macro level:} focuses on the overall logical trajectory and topical focus, judging whether the response matches the task and remains globally consistent.
(2) \textbf{Expositional level:} guided by information flow theory (Given$\rightarrow$New), evaluates whether the exposition unfolds naturally from shared context to new information and remains easy to follow.
(3) \textbf{Structural level:} inspects the completeness of the argumentation chain, including claims, evidence, reasoning, qualifiers, and consistency.

\medskip
\textbf{Eight Logical Dimensions and the Focus of the Rubric Question.}

\begin{enumerate}\setlength{\itemsep}{6pt}

\item \textbf{Task alignment \& claim clarity.} \\
\textbf{Definition:} Whether the response clearly addresses the core task of the QUERY in salient positions (e.g., the beginning or the end), and whether it states a clear, concentrated central claim. Evaluate whether the response stays on-topic or drifts. \\
\textbf{Question focus:} Compare which side more clearly responds to the question at the beginning or end, states a central stance, and maintains topic consistency.

\item \textbf{Global coherence.} \\
\textbf{Definition:} Whether the overall logical structure is clear and appropriate, and whether the content unfolds in a task-appropriate order (e.g., for analysis: background$\rightarrow$problem$\rightarrow$analysis$\rightarrow$conclusion; for comparison: comparison$\rightarrow$evaluation$\rightarrow$conclusion), forming a stable global thread. \\
\textbf{Question focus:} Compare which side is better organized overall, follows a more appropriate task logic, and whose conclusion better aligns with prior content.

\item \textbf{Internal consistency.} \\
\textbf{Definition:} Whether numbers, definitions, and stances remain consistent throughout, without ratio conflicts, shifting definitions, or self-contradictions. \\
\textbf{Question focus:} Compare which side is more internally consistent and logically self-contained.

\item \textbf{Concept introduction \& logical transition.} \\
\textbf{Definition:} The naturalness of information unfolding, especially the smoothness of the Given$\rightarrow$New transition, evaluated at two levels:
(a) Opening level: whether the beginning provides necessary background, context, or motivation before moving to the main claim. Jumping directly into abstract discussion or using undefined terms indicates weak transitions.
(b) Within-body level: when introducing new concepts, models, or terms, whether the response provides brief explanations or context so readers can understand their role and origin. \\
\textbf{Question focus:} Prioritize comparing which side better sets up background and naturally introduces the main topic at the opening; if both openings are reasonable, compare which side introduces new concepts more naturally and with sufficient explanation within the body.

\item \textbf{Local coherence.} \\
\textbf{Definition:} Whether adjacent paragraphs or sentences are logically connected, with reasonable transitions and continuity, rather than breaks or leaps. \\
\textbf{Question focus:} Compare which side has more reasonable paragraph-to-paragraph connections and more coherent transitions.

\item \textbf{Evidence sufficiency \& relevance.} \\
\textbf{Definition:} Whether key claims are supported by sufficient and on-topic evidence (e.g., data, experiments, facts, literature, cases), and whether the relation between evidence and conclusion is clear and traceable. \\
\textbf{Question focus:} Compare which side provides more concrete, query-relevant evidence that better supports the main conclusion.

\item \textbf{Warrants \& causal reasoning.} \\
\textbf{Definition:} Whether the response explains why the evidence supports the conclusion, whether the reasoning chain is complete, whether causal direction is reasonable, and whether there are logical jumps. \\
\textbf{Question focus:} Compare which side provides a more complete reasoning chain, clearer causal relations, and more justified step-by-step transitions.

\item \textbf{Qualifiers \& counterpoints.} \\
\textbf{Definition:} Whether the response identifies and handles potential counterexamples, uncertainty, or exceptions, and whether it avoids absolute claims by using appropriate qualifiers, concessions, or balanced framing. \\
\textbf{Question focus:} Compare which side better identifies counterexamples and sets conditions or assumptions that make the argument more balanced.

\end{enumerate}

\medskip
\textbf{Output Requirements.}
Output \textbf{only} a JSON array of length 8. Each object must correspond to one of Dimensions 1--8 in the same order.
Each object must contain exactly five fields:

\begin{itemize}\setlength{\itemsep}{3pt}
  \item ``aspect'': the exact name of one of the eight dimensions.
  \item ``question'': an instance-specific judgment question for the current QUERY and texts A/B.
  The question must be concrete and directly comparable between A and B, avoiding vague prompts such as ``Is it reasonable?'' or ``Is it sufficient?''.
  It should be phrased so annotators can decide which response is better for the specified logical focus.
  \item ``good\_example'': a positive example that matches the same question and illustrates what good performance looks like.
  \item ``bad\_example'': a negative example that matches the same question and illustrates what poor performance looks like.
  The good and bad examples must be paired to the same logical point under the same question.
  \item ``span\_hint'': a customized hint indicating linguistic or logical cues in TEXT A/B that annotators should attend to.
\end{itemize}

\medskip
\textbf{Generation Guidelines.}
\begin{enumerate}\setlength{\itemsep}{3pt}
  \item \textbf{Instance-specific.} All rubrics must be customized to the current QUERY and the content of TEXT A/B; avoid generic or template-like questions.
  \item \textbf{Comparative and self-contained.} Each ``question'' must be understandable on its own and explicitly invite an A-versus-B comparison.
  \item \textbf{Claim anchored.} Each ``question'' must name the concrete focal claim or topic in this instance (not vague phrases such as ``key claim'' or ``core point'').
  \item \textbf{Logic focus explicit.} Each ``question'' must specify the exact logical focus being compared (e.g., evidence relevance, warrant completeness, transition bridging), avoiding vague wording such as ``more reasonable'' or ``clearer'' without criteria.
  \item \textbf{Good/bad paired to the same point.} ``good\_example'' and ``bad\_example'' must correspond to the same rubric question and illustrate positive versus negative performance on that exact point.
  \item \textbf{Span hints tied to observable cues.} ``span\_hint'' must point to concrete linguistic or structural signals in TEXT A/B that annotators can inspect (e.g., explicit evidence statements, bridging sentences, qualifiers).
\end{enumerate}

\medskip
\textbf{Input.}\\
QUERY: \{QUERY\}\\
TEXT A: \{TEXT\_A\}\\
TEXT B: \{TEXT\_B\}

\medskip
\textbf{Output Format.}
The model must output a JSON array of length 8. Below is an example illustrating the required structure, keys, and style:

\smallskip
[
  \{
    ``aspect'': ``Evidence sufficiency \& relevance'',
    ``question'': ``Comparing A and B, which response provides more specific evidence (e.g., emissions data, country cases, or mechanism explanations) to support the claim that a carbon tax reduces emissions, and explains how this evidence justifies the conclusion?'',
    ``good\_example'': ``Cites concrete country-level data or empirical studies and explains how a carbon tax changes firm incentives, thereby reducing emissions.'',
    ``bad\_example'': ``States that a carbon tax is effective without providing concrete evidence or explaining the supporting mechanism.'',
    ``span\_hint'': ``Look for signals such as empirical data, study references, concrete cases, and explicit links between evidence and conclusions.''
  \},
  \ldots
]

\end{tcolorbox}

\captionof{figure}{Prompt used to generate context-aware, instance-specific rubric for eight-dimensional logical evaluation.}
\label{fig:rubric_prompt}
}
\twocolumn

\onecolumn
{
\small
\begin{tcolorbox}[
  enhanced, breakable,
  colback=softGray,
  colframe=badRed!60,
  title={LogicJudge Distillation Prompt},
  fonttitle=\bfseries,
  boxrule=0.6pt, arc=2pt, left=6pt, right=6pt, top=6pt, bottom=6pt]

\textbf{Prompt:}
You are a logic analysis expert tasked with comparing the logical quality of two long-form reports.
Given a user \textbf{QUERY}, two candidate reports (\textbf{A} and \textbf{B}), and a set of \textbf{context-aware logic rubrics} consisting of eight dimensions, you must determine which report exhibits stronger logical reasoning.

\vspace{6pt}
\textbf{Objective.}
Perform a rubric-guided \emph{pairwise preference judgment} over logical quality.
Your task is not to assess factual correctness or writing style, but to evaluate the \emph{logical structure and argumentative soundness} of the two reports.
You must first conduct explicit, dimension-level critiques as intermediate reasoning scaffolding, and then aggregate these judgments into an overall preference decision.

\vspace{8pt}
\textbf{Evaluation Procedure (STRICT).}
\begin{itemize}
  \item For each of the eight logic dimensions provided in the rubrics, determine whether report A or report B demonstrates stronger logical performance.
  \item Base each judgment on the dimension-specific comparison question, illustrative good/bad examples, and span-level reasoning cues provided in the rubric.
  \item For every dimension, produce (i) a winner label and (ii) a concise but concrete explanation grounded in logical reasoning.
  \item If both reports perform similarly well or similarly poorly on a dimension, this must be explicitly stated.
  \item After completing all dimension-level evaluations, synthesize them into a final overall preference decision.
\end{itemize}

\vspace{8pt}
\textbf{Evaluation Principles.}
\begin{itemize}
  \item Focus exclusively on logical reasoning quality, including claim clarity, evidence use, reasoning chains, coherence, and internal consistency.
  \item Do not rely on global impressions, fluency, verbosity, or stylistic features.
  \item Avoid vague judgments; all explanations must cite concrete logical characteristics.
  \item Do not quote or copy text from the reports; explanations must paraphrase logical patterns.
\end{itemize}

\vspace{8pt}
\textbf{Logic Dimensions (Fixed Schema).}
The evaluation must cover exactly the following eight dimensions:
\begin{enumerate}
  \item Task alignment \& claim clarity
  \item Global coherence
  \item Internal consistency
  \item Concept introduction \& logical transition
  \item Local coherence
  \item Evidence sufficiency \& relevance
  \item Warrants \& causal reasoning
  \item Qualifiers \& counterpoints
\end{enumerate}

\vspace{8pt}
\textbf{Input.}\\
QUERY: \{QUERY\}\\
TEXT A: \{TEXT\_A\}\\
TEXT B: \{TEXT\_B\}\\
Logic Rubric (JSON): {\{RUBRIC\_JSON\}}

\vspace{6pt}
\textbf{Output Format (STRICT).}
The model must output \emph{exactly one valid JSON object} following the schema below.
This structured format explicitly supervises both dimension-level reasoning and the final preference decision, enabling fine-grained analysis and scalable training.

\smallskip
\begin{lstlisting}
{
  "aspect_evaluations": {
    "logic_dimension_name": {
      "winner": "A>B" | "A<B" | "Tie",
      "explanation": "Explanation of logical differences"
    },
    ...
  },
  "overall_winner": "A>B" | "A<B" | "Tie",
  "overall_explanation": "A short paragraph summarizing the overall logical preference"
}
\end{lstlisting}

No additional text, commentary, or formatting outside this JSON object is allowed.

\end{tcolorbox}

\captionof{figure}{Prompt used to distill rubric-guided, dimension-level pairwise preferences for training LogicJudge.}
\label{fig:gen_train_prompt}
}
\twocolumn

\onecolumn
{
\small
\begin{tcolorbox}[
  enhanced, breakable,
  colback=softGray,
  colframe=badRed!60,
  title={LogicJudge Training Prompt},
  fonttitle=\bfseries,
  boxrule=0.6pt, arc=2pt, left=6pt, right=6pt, top=6pt, bottom=6pt]

\textbf{Prompt:}
You are an expert in logical analysis. Given a \textbf{QUERY}, two candidate reports (A and B), and a \textbf{context-aware rubric} with eight dimensions, you will decide which report is logically better on each dimension and provide targeted explanations. Finally, you will aggregate the eight dimensions into an overall verdict with a brief justification.

\vspace{6pt}
\textbf{Overall Objective.}
You must produce two levels of judgments:
(1) \textbf{Dimension-wise evaluation}: for each of the eight dimensions in the rubric, decide whether A or B is better and justify the decision;
(2) \textbf{Overall evaluation}: aggregate the eight dimension-wise decisions into a final overall winner (\texttt{overall\_winner}) and a short overall explanation.

\vspace{6pt}
\textbf{Procedure.}
(1) Interpret the rubric signals:
\texttt{question} specifies the logical focus of the dimension;
\texttt{good\_example} and \texttt{bad\_example} illustrate desired and undesired patterns;
\texttt{span\_hint} highlights cues and locations to inspect for reasoning links.
(2) Compare A vs. B for each dimension:
choose the winner and provide a concrete explanation (do not quote the original text verbatim);
if both are similarly strong or weak, state this explicitly.
(3) Produce the overall judgment:
\texttt{overall\_explanation} must be a coherent paragraph (no bullet points), stating the overall winner and the main reason.

\vspace{6pt}
\textbf{Evaluation Principles.}
Evaluate \textbf{logical structure and argumentation quality} only (not factual correctness or writing style).
All explanations must be grounded in the rubric focus and the reports' reasoning.
Avoid vague statements (e.g., ``better'' without justification).
Do not quote the reports verbatim; describe reasoning patterns in your own words.
The output must be a single JSON object with exactly the required fields and format.

\vspace{6pt}
\textbf{Eight Logic Dimensions (names must match exactly).}
\begin{enumerate}[leftmargin=1.4em, itemsep=0pt, topsep=2pt, parsep=0pt]
  \item Task alignment \& claim clarity
  \item Global coherence
  \item Internal consistency
  \item Concept introduction \& logical transition
  \item Local coherence
  \item Evidence sufficiency \& relevance
  \item Warrants \& causal reasoning
  \item Qualifiers \& counterpoints
\end{enumerate}

\vspace{6pt}
\textbf{Input Format.} \\
QUERY: \{QUERY\}\\
TEXT A: \{TEXT\_A\}\\
TEXT B: \{TEXT\_B\}\\
Logic Rubric (JSON): {\{RUBRIC\_JSON\}}

\vspace{6pt}
\textbf{Output Format (must follow strictly).}
\begin{lstlisting}
<think>
{
  "aspect_evaluations": {
    "Task alignment & claim clarity": {
      "winner": "A>B" | "A<B" | "both good" | "both bad",
      "explanation": "A concise, specific justification without verbatim quotes."
    },
    "Global coherence": {...},
    "Internal consistency": {...},
    "Concept introduction & logical transition": {...},
    "Local coherence": {...},
    "Evidence sufficiency & relevance": {...},
    "Warrants & causal reasoning": {...},
    "Qualifiers & counterpoints": {...}
  },
  "overall_explanation": "A single coherent paragraph summarizing the overall preference and main reasons (no lists)."
}
</think>
A>B
\end{lstlisting}

\vspace{4pt}
\textbf{Constraints.}
(1) The content inside \texttt{<think> ... </think>} must be a JSON object containing exactly \texttt{"aspect\_evaluations"} and \texttt{"overall\_explanation"}.
(2) \texttt{"aspect\_evaluations"} must cover all eight dimensions above. For each dimension, \texttt{"winner"} must be one of
\texttt{"A>B"}, \texttt{"A<B"}, \texttt{"both good"}, or \texttt{"both bad"}.
(3) After \texttt{</think>}, output exactly one final verdict token, chosen from
\texttt{"A>B"}, \texttt{"A<B"}, \texttt{"both good"}, or \texttt{"both bad"}, and output nothing else.

\end{tcolorbox}

\captionof{figure}{LogicJudge Training Prompt.}
\label{fig:logic_pref_prompt}
}
\twocolumn

\end{document}